\documentclass[sigconf, nonacm]{acmart}
\setcopyright{none}
\renewcommand\footnotetextcopyrightpermission[1]{}
\usepackage{multirow}
\usepackage{amsmath}
\usepackage{array}
\usepackage[table]{xcolor}
\usepackage{siunitx}
\definecolor{NYCband}{RGB}{248,252,255}
\definecolor{TKYband}{RGB}{250,248,255}
\definecolor{CAband}{RGB}{250,255,248}

\definecolor{FullRow}{RGB}{245,245,245}
\definecolor{BestCell}{RGB}{255,235,210}

\newcolumntype{Y}{>{\columncolor{NYCband}}S[table-format=1.4]}
\newcolumntype{T}{>{\columncolor{TKYband}}S[table-format=1.4]}
\newcolumntype{C}{>{\columncolor{CAband}}S[table-format=1.4]}

\newcommand{\bestab}[1]{%
  \multicolumn{1}{>{\columncolor{BestCell}\bfseries}S[table-format=1.4]}{#1}%
}

\newcommand{\rmv}{\textcolor{black!55}{\scriptsize($-$)}}

\definecolor{Top2Gray}{RGB}{90,90,90}
\definecolor{ImprovNum}{RGB}{220,20,120}
\newcommand{\best}[1]{{\bfseries\color{black}#1}}
\newcommand{\second}[1]{{\color{Top2Gray}\underline{#1}}}
\newcommand{\up}[1]{{\color{ImprovNum}\bfseries\small$\uparrow$\,#1}}

\usepackage{graphicx}
\usepackage[toc]{appendix}
\usepackage{subcaption}
\newcommand{\hNYC}[1]{\multicolumn{1}{>{\columncolor{NYCband}}c}{#1}}
\newcommand{\hTKY}[1]{\multicolumn{1}{>{\columncolor{TKYband}}c}{#1}}
\newcommand{\hCA}[1]{\multicolumn{1}{>{\columncolor{CAband}}c}{#1}}
\AtBeginDocument{%
 }

\setcopyright{acmlicensed}
\copyrightyear{2018}
\acmYear{2018}
\acmDOI{XXXXXXX.XXXXXXX}
\acmConference[Conference acronym 'XX]{Make sure to enter the correct
  conference title from your rights confirmation email}{June 03--05,
  2018}{Woodstock, NY}
\acmISBN{978-1-4503-XXXX-X/2018/06}

\begin{document}

\title{GTR-Mamba: Geometry-to-Tangent Routing Mamba for Hyperbolic POI Recommendation}

\author{Zhuoxuan Li}
\authornote{Both authors contributed equally to this research.}
\affiliation{
  \institution{College of Computer Science and Technology, Tongji University}
  \city{Shanghai}
  \country{China}
}
\email{li\_zhuoxuan@outlook.com}

\author{Jieyuan Pei}
\authornotemark[1]
\affiliation{
  \institution{College of Information Engineering, Zhejiang University of Technology}
  \city{Hangzhou}
  \country{China}
}
\email{peijieyuan@zjut.edu.cn}

\author{Tangwei Ye}
\affiliation{
  \institution{College of Computer Science and Technology, Tongji University}
  \city{Shanghai}
  \country{China}
}
\email{yetw@tongji.edu.cn}

\author{Zhongyuan Lai}
\affiliation{
  \institution{Shanghai Ballsnow Intelligent Technology Co., Ltd}
  \city{Shanghai}
  \country{China}
}
\email{zhongyuan.lai@ballsnow.com}

\author{Zihan Liu}
\affiliation{
  \institution{College of Computer Science and Technology, Tongji University}
  \city{Shanghai}
  \country{China}
}
\email{tongjilzh@gmail.com}

\author{Fengyuan Xu}
\affiliation{
  \institution{Hunan University}
  \city{Changsha}
  \country{China}
}
\email{xufengyuan126@gmail.com}

\author{Qi Zhang}
\affiliation{
  \institution{College of Computer Science and Technology, Tongji University}
  \city{Shanghai}
  \country{China}
}
\email{zhangqi\_cs@tongji.edu.cn}

\author{Liang Hu}
\authornote{Corresponding author}
\affiliation{
  \institution{College of Computer Science and Technology, Tongji University}
  \city{Shanghai}
  \country{China}
}
\email{rainmilk@gmail.com}

\renewcommand{\shortauthors}{Trovato et al.}

\begin{abstract}
Next Point-of-Interest (POI) recommendation is a critical task in modern Location-Based Social Networks (LBSNs), aiming to model the complex decision-making process of human mobility to provide personalized recommendations for a user's next check-in location. Existing hyperbolic POI recommendation models, predominantly based on rotations and graph representations, have been extensively investigated. Although hyperbolic geometry has proven superior in representing hierarchical data with low distortion, current hyperbolic sequence models typically rely on performing recurrence via expensive M\"obius operations directly on the manifold. This incurs prohibitive computational costs and numerical instability, rendering them ill-suited for trajectory modeling. To resolve this conflict between geometric representational power and sequential efficiency, we propose GTR-Mamba, a novel framework featuring Geometry-to-Tangent Routing. GTR-Mamba strategically routes complex state transitions to the computationally efficient Euclidean tangent space. Crucially, instead of a static approximation, we introduce a Parallel Transport (PT) mechanism that dynamically aligns tangent spaces along the trajectory. This ensures geometric consistency across recursive updates, effectively bridging the gap between the curved manifold and linear tangent operations. This process is orchestrated by an exogenous spatio-temporal channel, which explicitly modulates the SSM discretization parameters. Extensive experiments on three real-world datasets demonstrate that GTR-Mamba consistently outperforms state-of-the-art baselines in next POI recommendation.
\end{abstract}




\received{20 February 2007}
\received[revised]{12 March 2009}
\received[accepted]{5 June 2009}

\maketitle
\section{Introduction}

Point-of-Interest (POI) refers to a location that a user may find attractive or valuable. The proliferation of web-based location services and online social platforms \cite{sanchez2022point,wang2019sequential,yang2014modeling} has generated vast amounts of user-generated, geotagged content. This rich spatio-temporal data, such as check-ins and shared locations, makes it possible to predict places a user is likely to visit based on their preferences and contextual signals, fueling research on personalized next POI recommendation within the field of spatio-temporal data mining. This task is inherently challenging due to the complex interplay between users' hierarchical preferences and their dynamic, context-driven behaviors.

Existing recommendation systems are often based on sequential methods that personalize the processing of contextual information to better capture user preferences \cite{baral2018caps,baral2018close,wang2019spent}. Furthermore, given the powerful capability of Graph Neural Networks (GNNs) \cite{li2021you,qin2023disenpoi,xu2023revisiting} in integrating geographical information, they have been widely used in the task of next POI recommendation. Concurrently, recent breakthroughs in structured state-space sequence (S4) models have brought about significant efficiency improvements in sequential modeling. In particular, the Mamba architecture has gained significant prominence for its linear scaling capabilities \cite{chen2024geomamba,jiang2026hierarchical,qin2025geomamba}. However, these studies uniformly model user trajectories in Euclidean space, which struggle to effectively capture the inherent hierarchical and tree-like structures embedded in check-in behaviors.

\begin{figure}
\captionsetup{skip=3pt,belowskip=-15pt}
\centering
\includegraphics[width=\linewidth]{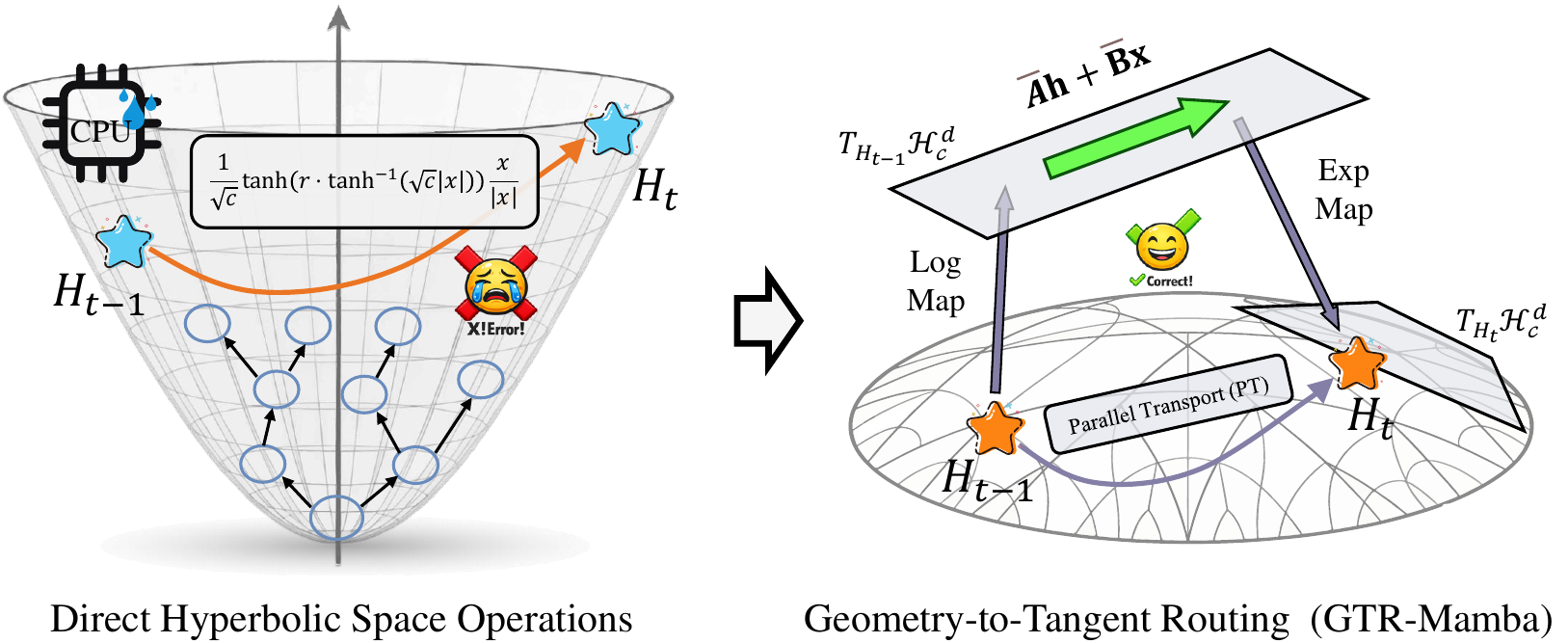}
\caption{(Left) Direct hyperbolic recurrence. (Right) Proposed \textbf{GTR-Mamba} with geometry-to-tangent routing.}
\label{task}
\end{figure}

In contrast, hyperbolic space naturally aligns with these latent hierarchies. Characterized by its negative curvature and exponential volume growth, it serves as an ideal inductive bias for modeling power-law distributions \cite{peng2021hyperbolic,yang2022hyperbolic}. In POI scenarios, hyperbolic embeddings can effectively organize multi-granular semantics (e.g., regions, categories, and POIs) and capture long-tail relationships within a low-dimensional space. Existing literature has explored embeddings in hyperbolic space \cite{feng2020hme,liu2025hyperbolic,qiao2025hyperbolic}, establishing it as a natural choice for modeling hierarchical data. These studies, often integrating hyperbolic geometry with rotation-based methods or variational graph auto-encoders, have demonstrated promising results. However, a critical gap remains in Sequential Modeling within current hyperbolic recommendation frameworks. Most existing works focus predominantly on static graph representations. The few attempts to construct hyperbolic sequence models typically rely on performing heavy M\"obius operations directly on the manifold. This prohibitive computational cost and potential numerical instability severely constrain their scalability in scenarios involving long trajectories and large-scale candidate sets.

To address this fundamental conflict between the representational benefits of hyperbolic geometry and the efficiency of sequence modeling, we propose GTR-Mamba, a novel framework featuring Geometry-to-Tangent Routing. As illustrated in Figure \ref{task}, we design a novel Mamba layer that executes state updates within the computationally efficient Euclidean tangent space. Crucially, to mitigate the precision loss caused by tangent space approximation, we introduce a Parallel Transport (PT) mechanism. As the reference point changes dynamically, this mechanism rigorously transports the hidden vector from the previous tangent space to the current one, thereby maintaining geometric consistency across recursive steps. Furthermore, the discretization and update of the internal State Space Model (SSM) are explicitly driven by exogenous spatio-temporal conditions to adapt to irregular time intervals and contextual shifts. Overall, GTR-Mamba retains the superiority of hyperbolic representation while balancing efficiency and training stability in long-trajectory scenarios.

Our contributions are summarized as follows:
\begin{itemize}
    \item We propose a novel hyperbolic Mamba layer that routes complex dynamic sequence updates to the computationally stable and efficient Euclidean tangent space for execution, while ensuring geometric fidelity via a Parallel Transport mechanism.
    \item We propose a context-explicit driven variable-step selective SSM, where internal dynamic state transitions adaptively adjust based on external spatiotemporal signals to address complex temporal and contextual shifts.
    \item Extensive experiments were conducted on three real-world LBSN datasets. The results confirm that our proposed GTR-Mamba model demonstrates superior overall performance compared to state-of-the-art baseline methods.
\end{itemize}


\section{Related Work}
\label{sec:related_work}
\subsection{Next POI Recommendation}
Next Point-of-Interest (POI) recommendation often relies on modeling the complex transitional and sequential patterns within users' historical check-ins. Leveraging the powerful deep modeling capabilities of deep neural networks, sequential models such as LSTM or RNN have been employed to treat the POI task as a sequence prediction problem \cite{feng2018deepmove,liu2016predicting,wang2021reinforced,wu2020personalized}. Concurrently, variants of the attention mechanism \cite{duan2023clsprec,luo2021stan,xue2021mobtcast,zhang2022next} have been widely adopted due to their ability to focus on more critical parts of historical spatio-temporal information, thereby integrating richer contextual representations. Graph Neural Networks (GNNs) \cite{li2021you,qin2023disenpoi,wang2022learning,wang2023adaptive,yan2023spatio,xu2023revisiting} have also achieved significant success by further modeling geographical dependencies through neighborhood aggregation. Notably, some research has already recognized the importance of hierarchical structures for the POI recommendation task, including the introduction of auxiliary information such as POI categories \cite{yu2020category,zang2021cha,zhang2020modeling} and geographical regions \cite{lian2020geography,lim2022hierarchical,xie2023hierarchical} to enhance recommendation performance. Furthermore, tree-based methods \cite{baral2018caps,chen2023dynamic,huang2024learning,lu2020glr} have also been proposed, as trees inherently possess a hierarchical structure. 

Owing to Mamba's formidable long-sequence modeling capabilities, several Mamba-based methods have recently been introduced. For instance, Chen et al. \cite{chen2024geomamba} leverage a combination of hierarchical geographical encoding and Mamba to achieve awareness of geographical sequences, while Qin et al. \cite{qin2025geomamba} utilize Mamba and the GaPPO operator to extend the state space for modeling multi-granularity spatio-temporal transitions. Although these recommendation methods have achieved excellent results, these state-of-the-art models predominantly operate in Euclidean space, which inherently struggles to preserve the underlying hierarchical and tree-like structures of human mobility.

\subsection{Hyperbolic Representation Learning}

Hyperbolic geometry has been extensively adopted in recommendation systems due to its superior capacity for modeling hierarchical structures \cite{sun2021hgcf}. Its applications span diverse domains, including knowledge-aware \cite{chen2022modeling,du2022hakg}, social \cite{wang2021hypersorec,yang2023hyperbolic}, session-based \cite{guo2023hyperbolic}, and news recommendation \cite{wang2023hdnr}. Notably, hyperbolic collaborative filtering \cite{li2022hyperbolic,yang2022hicf} has demonstrated significant improvements over Euclidean baselines. In POI recommendation, HME \cite{feng2020hme} embeds interactions into a Poincaré ball, while recent works like HMST \cite{qiao2025hyperbolic} and HVGAE \cite{liu2025hyperbolic} further advance this by utilizing hyperbolic rotations and variational graph autoencoders to capture multi-semantic transitions. However, these methods largely focus on static node representation or graph structures, often neglecting the efficient modeling of long-term sequential evolution.

To address temporal dynamics, recent research has integrated hyperbolic geometry into sequence encoders. Early attempts like HVACF \cite{shimizu2024fashion} and the work by Patil et al. \cite{patil2025hierarchical} introduced hyperbolic manifolds into user-item representation learning and pooled feature projection. To handle longer sequences, Hypformer \cite{yang2024hypformer} adapts Transformer attention to Lorentz space, though it retains high computational complexity. More recently, geometry-aware methods like HMamba \cite{zhang2025hmamba} and Hybrid-Emba3D \cite{liu2025hybrid} have explored linear-complexity architectures. Nevertheless, existing hyperbolic SSMs typically rely on computationally intensive Möbius operations for state updates, leading to higher per-step overhead in practice.

\section{Preliminary}
\label{sec:p}
\subsection{Basic Definition}

Let $\mathcal{U}$, $\mathcal{P}$, $\mathcal{C}$, and $\mathcal{R}$ be the sets of users, POIs, categories, and regions, respectively, where $|\mathcal{P}|$ is the total number of POIs. Each POI is associated with location information, represented by geographical coordinates, and category information that reflects its function. The regions are constructed by partitioning the entire geographical area based on the collected coordinates, which determines the region to which each POI belongs.

A check-in, denoted as $s = (u, p, t, c, r)$, records the event of a user $u \in \mathcal{U}$ visiting a specific POI $p \in \mathcal{P}$ at a timestamp $t$. Here, $c \in \mathcal{C}$ and $r \in \mathcal{R}$ represent the category and region of POI $p$, respectively. We represent the check-in sequence of a user $u$ as $\mathcal{S}_u = \{s_1, s_2, \dots, s_{l_u}\}$, where $s_i$ is the $i$-th check-in of user $u$, and $l_u$ is the length of the sequence $\mathcal{S}_u$.

Given a user's historical check-in sequence $\mathcal{S}_u$, the objective of next POI recommendation is to predict the POI $p_{l_u+1}$ that the user $u$ is most likely to visit next.

\subsection{Hyperbolic Space}

Let $\mathbb{H}^d_c$ denote a $d$-dimensional hyperbolic space with negative curvature $-1/c<0$, where $c>0$. In this paper, we adopt the Lorentz model embedded in $\mathbb{R}^{d+1}$. The $d$-dimensional Lorentz model is defined as a Riemannian manifold with constant negative curvature $-1/c$: $\mathbb{L}_c^d=(\mathbb{H}_c^d,g_x^c)$, where $\mathbb{H}_c^d=\{\,x\in\mathbb{R}^{d+1}:\langle x,x\rangle_L=-c;\ 
x_0>0\,\}$, $g_x^c(u,v)=\langle u,v\rangle_L.$
We adopt the convention where the time coordinate is first: $x=(x_0,x_1,\dots,x_d)$, with $x_0$ being the temporal component. The Lorentzian inner product $\langle\cdot,\cdot\rangle_L$ is given by $\langle x,y\rangle_L=-x_0y_0+\sum_{i=1}^{d}x_iy_i$. 
The commonly used squared Lorentz distance \cite{law2019lorentzian} is defined as:
\begin{equation}
    d_L^2(x,y):=-2c-2\langle x,y\rangle_L.
\end{equation}
This distance metric captures the hyperbolic geometry and is effective for representing hierarchical relationships.

\paragraph{Tangent space and base-point maps.}
For any point $x\in\mathcal{H}_c^d$, there exists a $d$-dimensional vector space $T_x\mathbb{H}_c^d$, known as the tangent space at $x$ \cite{peng2021hyperbolic}. The exponential map $\exp_x(\cdot):T_x\mathbb{H}_c^d\to\mathcal{H}_c^d$ and logarithmic map $\log_x(\cdot):\mathcal{H}_c^d\to T_x\mathbb{H}_c^d$ transform between the tangent space and the manifold \cite{ganea2018hyperbolic}. Importantly, the subscript in $\exp_x(\cdot)$ and $\log_x(\cdot)$ denotes the \emph{base point} (i.e., the reference point defining the tangent space). Let the origin $o\in\mathcal{H}_c^d$ be defined as $o=(\sqrt{c},0,\ldots,0)$.
Let $\|v\|_L=\sqrt{\langle v,v\rangle_L}$ (in the tangent space, this norm is real and non-negative). The exponential and logarithmic maps based at the origin $o$ are then given by:
\begin{equation}
    \exp_o(v)=\cosh\!\left(\frac{\|v\|_L}{\sqrt{c}}\right)o+\sqrt{c}\sinh\!\left(\frac{\|v\|_L}{\sqrt{c}}\right)\frac{v}{\|v\|_L},
\end{equation}
\begin{equation}
    \log_o(x)=\frac{\operatorname{arccosh}\!\left(-\frac{\langle o,x\rangle_L}{c}\right)}{\left\|x+\frac{\langle o,x\rangle_L}{c}o\right\|_L}\left(x+\frac{\langle o,x\rangle_L}{c}o\right).
\end{equation}

\paragraph{Base-point maps and parallel transport.}
The subscript in $\exp_x(\cdot)$ and $\log_x(\cdot)$ denotes the \emph{base point} $x$ that defines the tangent space $T_x\mathbb{H}_c^d$:
$\exp_x: T_x\mathbb{H}_c^d \to \mathbb{H}_c^d$ and $\log_x: \mathbb{H}_c^d \to T_x\mathbb{H}_c^d$.

\paragraph{Parallel transport (PT)}
Parallel transport moves a tangent vector from one tangent space to another along the (unique) geodesic \cite{bose2020latent}.
Formally, for a geodesic $\gamma$ with $\gamma(0)=x,\gamma(1)=y$, PT is defined by the Levi--Civita connection:
\[
\nabla_{\dot\gamma(t)} v(t)=0,\quad v(0)=v_x,\quad \mathrm{PT}_{x\to y}(v_x)=v(1),
\]
which preserves the Riemannian inner product (lengths and angles).

In the Lorentz model $\mathcal{H}_c^d=\{x:\langle x,x\rangle_L=-c,\ x_0>0\}$ with metric $g_x^c(u,v)=\langle u,v\rangle_L$,
a closed-form expression for PT along the geodesic from $x$ to $y$ is:
\begin{equation}
\mathrm{PT}_{x\rightarrow y}(v)
= v + \frac{\langle y, v\rangle_L}{c-\langle x, y\rangle_L}\,(x+y),
\qquad v\in T_x\mathbb{H}_c^d.
\end{equation}
This mapping satisfies:
\begin{equation}
    \mathrm{PT}_{x\to y}(v)\in T_y\mathbb{H}_c^d,\quad
\langle \mathrm{PT}_{x\to y}(v_1), \mathrm{PT}_{x\to y}(v_2)\rangle_L
= \langle v_1, v_2\rangle_L.
\end{equation}

In the subsequent derivations and implementation, we set the curvature parameter $c=1$. Following prior work \cite{chami2019hyperbolic,liu2025hyperbolic}, we consistently use $o=(1,0,\ldots,0)$ as the reference point and employ $\exp_o(\cdot)$ and $\log_o(\cdot)$ for transformations between $\mathbb{H}^d$ and the tangent space $T_o\mathbb{H}^d$ when a fixed base point is sufficient; in the moving-base formulation, we instead use $\exp_x(\cdot)$, $\log_x(\cdot)$, and $\mathrm{PT}_{x\rightarrow y}(\cdot)$ with dynamically changing base points.

\section{The Proposed Model}
\label{sec:m}

\begin{figure*}
\captionsetup{skip=2pt,belowskip=-5pt}
 \centering
 \includegraphics[width=\textwidth]{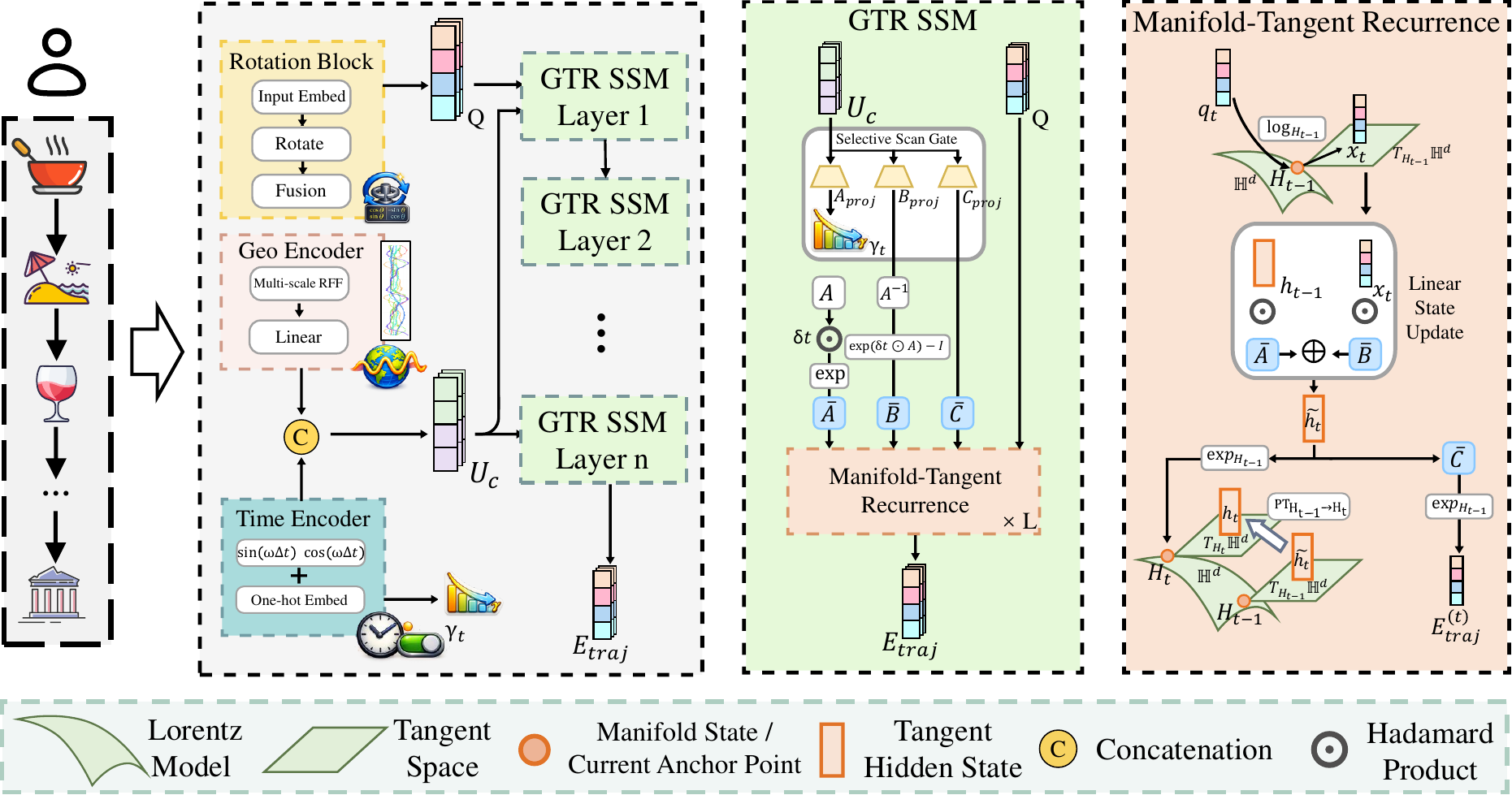}
\caption{The overall architecture of GTR-Mamba. The framework comprises three key components: (1) \textit{Hyperbolic Rotational Alignment} via rotation-based contrastive learning; (2) An exogenous \textit{Spatio-Temporal Channel} encoding Euclidean contexts; and (3) The \textit{GTR-Mamba Layer}, which routes state transitions to the tangent space using Parallel Transport (PT) while modulating SSM parameters via external context signals.}
\label{GTR}
 \Description[none]{}
\end{figure*}

\subsection{Overview}
As illustrated in Figure \ref{GTR}, the overall framework of our proposed GTR-Mamba consists of three integral components. We learn hyperbolic embeddings and relation-specific rotations jointly with the sequential model, and use them to form the trajectory input representation. Parallel to this, we employ a Euclidean Spatio-Temporal Channel that encodes geographical contexts using spherical multi-scale Random Fourier Features (RFF), while capturing temporal dynamics via multi-frequency sine-cosine encoding. Subsequently, the sequence representation is routed into the core GTR-Mamba Layer, where the internal State Space Model (SSM) is explicitly driven by the encoded Euclidean spatio-temporal signals. Finally, the next POI is predicted through a dual-path scoring mechanism that aggregates scores from both the hyperbolic manifold and the Euclidean tangent space.

\subsection{Hyperbolic Rotational Alignment}
\label{emb}
Following prior research \cite{qiao2025hyperbolic, feng2020hme}, we optimize entity representations on the Lorentz manifold to capture static hierarchical and transitional relationships. We construct a multi-relational graph from historical check-ins, encompassing User-POI interactions, sequential POI transitions (within a 6-hour window), and their corresponding category and region transitions. To align diverse semantics, learnable isometric rotations $Rot(\cdot)$ are applied to source embeddings to better align them with targets while preserving the Lorentzian inner product. The model is optimized using a contrastive loss that maximizes the similarity scores $s_t$ (including positive sample score $\text{s}_{\text{pos}, t}$ and negative sample score $\text{s}_{\text{neg}, t}$) for observed edges across all types $t \in \{up, pp, cc, rr\}$:
\begin{equation}\mathcal{L}_{\text{rot}} = -\sum_{t} \left( \sum \log \sigma(\text{s}_{\text{pos}, t}) + \sum \log \sigma(-\text{s}_{\text{neg}, t}) \right),
\end{equation}
where $\sigma(\cdot)$ is the sigmoid function.

Given the hyperbolic embeddings $\mathbf{E}_p, \mathbf{E}_c, \mathbf{E}_u, \mathbf{E}_r$, we fuse these multi-modal representations via semantic composition in the tangent space. Specifically, we project each entity's embedding onto the tangent space at the origin $o$ using the logarithmic map and compute the sequence representation $\mathbf{q}_t$:
\begin{equation}
\mathbf{q}_{\text{t}} = \alpha_u \log_o(\mathbf{E}_u) + \alpha_p \log_o(\mathbf{E}_p) + \alpha_c \log_o(\mathbf{E}_c) + \alpha_r\log_o(\mathbf{E}_r),
\end{equation}
where $\alpha$ are hyperparameters used to prioritize different semantic objectives. This composite vector is finally mapped back to the manifold to obtain the hyperbolic sequence representation $\mathbf{q}_t\in\mathbb{H}^d$. We then stack $\{\mathbf{q}_t\}_{t=1}^L$ to form the initialized trajectory $\mathbf{Q}=[\mathbf{q}_1,\ldots,\mathbf{q}_L]\in\mathbb{H}^{L\times d}$ for subsequent sequence modeling.

\subsection{Spatio-temporal Channel}

While hyperbolic embeddings capture hierarchical semantics, they lack explicit awareness of linear spatio-temporal dynamics (e.g., continuous time intervals). Instead of forcing Euclidean spatio-temporal signals (e.g., time intervals, grid coordinates) into the hyperbolic manifold—which risks geometric distortion and semantic misalignment—we encode them via an exogenous Euclidean channel to linearly modulate the SSM dynamics.

\subsubsection{Geographic Embedding Module}
Our geographical embedding module encodes latitude and longitude via multi-scale Random Fourier Features (RFF) to obtain a high-dimensional harmonic representation of spatial contexts.

First, we map latitude and longitude to unit vectors on a sphere. We then sample a multi-scale Gaussian frequency matrix and compute harmonic features by sine-cosine projections, yielding global multi-scale geographical features $\mathbf{E}_{\text{RFF}}$. Finally, a linear projection is applied to obtain the geographic embedding $\mathbf{E}_g \in \mathbb{R}^{L \times d_{\text{geo}}}$.

\subsubsection{Temporal Feature Module}
The temporal feature module is responsible for generating temporal representations and a decay factor, which are used to modulate the dynamic evolution of the SSM.

We first extract the time interval $\Delta t \in \mathbb{R}^{L \times 1}$, the day of the week ($\text{dow}$), and the hour of the day ($\text{hour}$) from the time series. These are then fused into a feature vector:
\begin{equation}
    \mathbf{E}_t = \text{Concat}[\Delta t; \sin(\omega \Delta t); \cos(\omega \Delta t); \text{OH}(\text{dow}); \text{OH}(\text{hour})],
\end{equation}
where $\omega \in \mathbb{R}^{M}$ denotes a vector of $M$ logarithmically spaced frequencies, $\mathbf{E}_{t} \in \mathbb{R}^{L \times d_{\text{time}}}$, and $\text{OH}(\text{dow})$ and $\text{OH}(\text{hour})$ are 7-dimensional and 24-dimensional one-hot encodings, respectively. This design captures both the periodicity and long-term trends of temporal data.

The feature vector is then projected to a dimension of $d_{\text{time}}$, and a decay factor is computed via a gated mechanism:
\begin{equation}
    \gamma_t = \sigma(\mathbf{E}_t \cdot \mathbf{w}_{\text{gate}}),
\end{equation}
where $\gamma_t \in \mathbb{R}^{L \times 1}$ and $\mathbf{w}_{\text{gate}} \in \mathbb{R}^{d_{\text{time}}}$ is a learnable weight vector. This decay factor $\gamma_t$ modulates the step size of the SSM, simulating the influence of time intervals on the trajectory dynamics. Let $\mathbf{u}_c = \text{Concat}[\mathbf{E}_g ; \mathbf{E}_t]$ denote the combined Euclidean context feature.

\subsection{GTR-Mamba Layer}

To address the modeling challenges of sequence-encoded tasks, we adopt the Mamba framework \cite{gu2024mamba}. Its selective scanning mechanism, enabled by dynamic step sizes and input gating, adaptively captures variations in temporal intervals and external spatiotemporal contexts. Furthermore, Mamba's linear recursive computation facilitates efficient dynamic updates in a local tangent space, making it particularly suitable for trajectory modeling on non-Euclidean manifolds.

For enhanced stability and computational efficiency, we employ a fixed diagonal matrix:
\begin{equation}
\mathbf{A} = -\text{diag}(\log(2), \log(3), \dots, \log(d+1)).
\end{equation}

This structure naturally accommodates the multi-time-scale characteristics of trajectory data, ranging from short-term frequent check-ins to long-term behavioral patterns. To address the lack of cross-channel mixing in the diagonal matrix $\mathbf{A}$, which lacks cross-channel coupling, we implement dynamic channel modulation on the input side through context-driven selective gating.

The step size $\delta t$ is dynamically generated based on the Euclidean features $\mathbf{u}_c$:
\begin{equation}
\delta t = (A_{\text{proj}}(\mathbf{u}_c) \cdot \mathbf{W}_{dt} + \mathbf{b}_{dt}) \odot \gamma_t,
\end{equation}
where $A_{\text{proj}}: \mathbb{R}^d \to \mathbb{R}^d$, $\mathbf{W}_{dt} \in \mathbb{R}^{d}$ is a learnable vector, $\mathbf{b}_{dt}\in\mathbb{R}^{d}$ is a learnable bias, and $\gamma_t \in \mathbb{R}^{L\times 1}$ is the decay factor obtained from the temporal encoding module.

Subsequently, we discretize the continuous SSM. The state transition matrix $\bar{\mathbf{A}} \in \mathbb{R}^{L \times d}$ is:
\begin{equation}
\bar{\mathbf{A}} = \exp(\delta t \odot \mathbf{A}),
\end{equation}
and the input matrix $\bar{\mathbf{B}} \in \mathbb{R}^{L\times d}$ is:
\begin{equation}
\bar{\mathbf{B}} = (\exp(\delta t \odot \mathbf{A}) - \mathbf{I}) \odot \mathbf{A}^{-1},
\end{equation}
where $\mathbf{I}$ is the identity matrix.

To inject the exogenous contextual conditions, the input matrix $\bar{\mathbf{B}}$ is modulated by selective weights derived from the Euclidean context features:
\begin{equation}
\bar{\mathbf{B}} \leftarrow \bar{\mathbf{B}} \odot B_{\text{proj}}(\mathbf{u}_c),
\end{equation}
where $B_{\text{proj}}: \mathbb{R}^d \to \mathbb{R}^d$. This element-wise multiplication allows each state dimension to independently determine its input strength based on the spatio-temporal context.

Prior research has demonstrated that tangent space approximation introduces distortion, and restricting operations to a fixed tangent space limits the model's ability to leverage the rich information of the manifold \cite{zhang2004principal,bose2020latent}. Inspired by Bose et al., our model incorporates Parallel Transport (PT) to enable selective scanning within a moving tangent space anchored at the current manifold state \cite{bose2020latent}. Parallel transport serves as the canonical generalization of Euclidean translation to Riemannian manifolds. Given that the core recurrence of Mamba involves linear state transitions, PT provides the necessary geometric operator to transport updated hidden states between varying reference points while preserving the Riemannian metric structure.

Let $\mathbf{H}_{t-1}\in\mathbb{H}^d$ denote the manifold anchor at step $t-1$. We first map the current hyperbolic input $\mathbf{q}_t\in\mathbb{H}^d$ to the local tangent space at $\mathbf{H}_{t-1}$:
\begin{equation}
\mathbf{x}_t = \log_{\mathbf{H}_{t-1}}(\mathbf{q}_t)\in T_{\mathbf{H}_{t-1}}\mathbb{H}^d.
\end{equation}

We maintain a tangent-space hidden state $\mathbf{h}_{t-1}\in T_{\mathbf{H}_{t-1}}\mathbb{H}^d$ and update it linearly in the same local tangent space:
\begin{equation}
\tilde{\mathbf{h}}_t = \bar{\mathbf{A}}_t \odot \mathbf{h}_{t-1} + \bar{\mathbf{B}}_t \odot \mathbf{x}_t.
\label{eq:linear_recurrence}
\end{equation}
Note that $\tilde{\mathbf{h}}_t$ resides in the tangent space of the previous anchor $\mathbf{H}_{t-1}$.

We then map the updated tangent vector $\tilde{\mathbf{h}}_t$ back to the manifold via the exponential map anchored at $\mathbf{H}_{t-1}$:
\begin{equation}
\mathbf{H}_t = \exp_{\mathbf{H}_{t-1}}(\tilde{\mathbf{h}}_t)\in\mathbb{H}^d.
\end{equation}


Formally, the validity of the GTR mechanism is grounded in the differential geometric principle that linear operations are strictly local. While the SSM core performs efficient linear recurrence, it implicitly assumes a flat vector space structure. On a curved Lorentz manifold, however, the tangent spaces $T_{\mathbf{H}_{t-1}}\mathbb{H}^d$ and $T_{\mathbf{H}_t}\mathbb{H}^d$ are distinct vector spaces. Unlike in Euclidean space, where vectors can be trivially translated due to zero curvature, in hyperbolic space, the basis vectors of the tangent space undergo rotation as they move along the manifold---a phenomenon governed by the manifold's non-zero curvature and holonomy. Consequently, directly reusing a hidden vector from the previous step without adjustment is mathematically ill-posed, as it ignores the geometric twisting of the coordinate system.

To rigorously bridge this gap, we employ the Levi-Civita Parallel Transport (PT) along the unique geodesic connecting the consecutive anchors $\mathbf{H}_{t-1}$ and $\mathbf{H}_t$. This operator defines an isometric isomorphism between the disjoint tangent spaces, preserving the Riemannian metric properties (i.e., inner products, norms, and angles). Conceptually, this step acts as a discrete realization of \textit{covariant transport}, explicitly rectifying the frame misalignment caused by the manifold's curvature. By compensating for the geometric rotation, PT ensures that the historical state $\tilde{\mathbf{h}}_t$ is faithfully translated into the new reference frame $T_{\mathbf{H}_t}\mathbb{H}^d$, maintaining the physical consistency of the sequence memory across recursive updates:
\begin{equation}
\mathbf{h}_t = \mathrm{PT}_{\mathbf{H}_{t-1}\rightarrow \mathbf{H}_t}(\tilde{\mathbf{h}}_t)\in T_{\mathbf{H}_t}\mathbb{H}^d.
\end{equation}
This transported vector $\mathbf{h}_t$ effectively becomes the valid hidden state for the subsequent time step.

Then we introduce a context-driven $C$-projection. Specifically, for each time step we generate a vector $\mathbf{c}_t\in\mathbb{R}^{d}$ from the Euclidean spatio-temporal feature $\mathbf{u}_c$ (time-step sliced), and apply an element-wise product along the state dimension to modulate the recurrent state:
\begin{equation}
\mathbf{c}_t = \sigma(C_{\text{proj}}(\mathbf{u}_c))\in\mathbb{R}^{d},\qquad
\mathbf{o}_t = \mathbf{c}_t \odot \tilde{\mathbf{h}}_t \in T_{\mathbf{H}_{t-1}}\mathbb{H}^d,
\end{equation}
where $C_{\text{proj}}:\mathbb{R}^d\to\mathbb{R}^d$. The per-step trajectory embedding is obtained by mapping the readout vector back to the manifold:
\begin{equation}
\mathbf{E}_{\mathrm{traj}}^{(t)} = \exp_{\mathbf{H}_{t-1}}(\mathbf{o}_t)\in\mathbb{H}^{d}.
\end{equation}

By stacking the per-step embeddings over $t=1,\ldots,L$, we obtain $\mathbf{E}_{\mathrm{traj}}=[\mathbf{E}_{\mathrm{traj}}^{(1)},\ldots,\mathbf{E}_{\mathrm{traj}}^{(L)}]\in\mathbb{H}^{L\times d}$. 

Compared to a fixed-origin global linearization, the moving-base formulation performs local linearization along the trajectory and propagates memory with parallel transport, reducing accumulated distortion on long and irregular trajectories.

\subsection{Prediction and Loss}

Having encoded the user's dynamic preference evolution into the trajectory embedding $\mathbf{E}_{\mathrm{traj}}$, the final step is to predict the next likely visit. Our method also forecasts potential transitions between categories and regions to aid in the next location prediction. In other words, we integrate the results from these multiple tasks to formulate the final recommendation. Here, we use POI prediction as an illustrative example; the prediction process for the other tasks is analogous. The total prediction loss is the sum of these individual losses:
\begin{equation}
    \mathcal{L}_{\text{all}} = \mathcal{L}_{\text{poi}} + \mathcal{L}_{\text{cat}} + \mathcal{L}_{\text{reg}}+\mathcal{L}_{\text{rot}}.
\end{equation}

Our prediction component performs scoring from both the hyperbolic and tangent spaces.

To capture the geometric relationships between embeddings, we compute the squared Lorentz distance, $d_L^2(\cdot, \cdot)$, on the manifold between the GTR-Mamba output trajectory embedding, $\mathbf{E}_\mathrm{traj}$, and all candidate entities (POIs, categories, or regions). The similarity score is defined as:
\begin{equation}
    \text{s}_{\text{hyperbolic}} = -\frac{\sqrt{d_L^2(\mathbf{E}_\mathrm{traj}, \text{p})}}{\tau},
\end{equation}
where $\tau$ is a learnable temperature parameter. For POI prediction, $\text{p}$ represents the embeddings of all POIs $\mathbf{E}_{\mathcal{P}} \in \mathbb{H}^{|\mathcal{P}| \times d}$. This distance-based score leverages the geometric properties of the Lorentz manifold by directly comparing embeddings in the hyperbolic space, which is well-suited for hierarchical data.

Concurrently, the tangent vector of the trajectory is decoded through a linear layer to produce logit scores for the candidate entities:
\begin{equation}
    \text{s}_{\text{tangent}} = \text{Linear}(\log_o(\mathbf{E}_\mathrm{traj})).
\end{equation}
This provides a direct classification prediction that captures the linear patterns within the tangent space.

To balance the geometric scores and the linear predictions, we introduce a learnable mixing parameter $\beta$. The final prediction score is a weighted combination:
\begin{equation}
    \text{s} = \beta \cdot \text{s}_{\text{tangent}} + (1 - \beta) \cdot \text{s}_{\text{hyperbolic}}.
\end{equation}
This formulation combines the hierarchical expressive power of hyperbolic distance with the flexibility of a linear decoder, adaptively adjusting the weights of the two components via the parameter $\beta$.

For the POI, category, and region prediction tasks, we employ the cross-entropy loss, yielding the respective losses $\mathcal{L}_{\text{poi}}$, $\mathcal{L}_{\text{cat}}$, and $\mathcal{L}_{\text{reg}}$. 

\section{Experiments and Results}
\label{sec:e}

\subsection{Dataset}
We evaluate our proposed model on three datasets collected from two real-world check-in platforms: Foursquare \cite{yang2014modeling} and Gowalla \cite{yuan2013time}. The Foursquare dataset includes two subsets, which are collected from New York City (NYC) in the USA and Tokyo (TKY) in Japan. The Gowalla dataset includes one subset collected from California and Nevada (CA). Their detailed statistics are in Table \ref{tab:data_summary}. The density is calculated as the total number of visits divided by (number of users × number of POIs), which is used to reflect the sparsity level between users and POIs.
Following prior work \cite{sun2020go}, we remove POIs with fewer than five check-ins, segment each user's trajectory into sequences of length 3--101, and split the data into training, validation, and test sets in an 8:1:1 ratio.
\begin{table}[ht]
\captionsetup{aboveskip=0pt, belowskip=3pt}
\centering
\caption{Statistics of the evaluated datasets.}
\renewcommand{\arraystretch}{1.2}
\begin{tabular}{lcccccc}
    \toprule
    & User & POI & Category & Traj & Check-in & Density \\
    \midrule
    \rowcolor{NYCband} NYC & 1,047 & 4,980 & 318 & 13,955 & 101,760 & 0.016 \\
    \rowcolor{TKYband} TKY & 2,281 & 7,832 & 290 & 65,914 & 403,148 & 0.018 \\
    \rowcolor{CAband}  CA  & 3,956 & 9,689 & 296 & 42,982 & 221,717 & 0.005 \\
    \bottomrule
\end{tabular}%
\label{tab:data_summary}
\end{table}

\subsection{Experiment Setting}
In our experiments, the curvature parameter, c, of the hyperbolic space was set to 1. We configured the model with the following default hyperparameters: a batch size of 128, a learning rate of 0.001, and a training duration of 50 epochs. Weights $\alpha_u, \alpha_p, \alpha_c $ and $ \alpha_r$ were set to 0.5, 0.3, 0.1 and 0.1 respectively. The entire geographical area was partitioned into 40 regions. The embedding dimension, d, was set to 64, with the geographical and temporal encoding dimensions set to 16 and 24, respectively. For the auxiliary relational alignment loss, we used 5 negative samples per positive instance. The Mamba architecture consisted of 2 layers.

We evaluated the recommendation performance using Top-k Normalized Discounted Cumulative Gain (NDCG@k) and Mean Reciprocal Rank (MRR). We report results for $k \in \{1, 5, 10\}$. The model is implemented in PyTorch and executed on an NVIDIA GeForce RTX 4090 GPU.

\begin{table*}[t]
\captionsetup{aboveskip=0pt, belowskip=3pt}
\centering
\caption{Overall performance comparison with baseline models on three datasets.
\textbf{Top\mbox{-}1},\; \textcolor{Top2Gray}{\underline{Top\mbox{-}2}}.}
\setlength{\tabcolsep}{2.5pt} 
\renewcommand{\arraystretch}{1.15}
\resizebox{\textwidth}{!}{%
    \begin{tabular}{l *{4}{Y} *{4}{T} *{4}{C}}
    \toprule
    \multirow{2.5}{*}{\textbf{Method}} &
    \multicolumn{4}{c}{\cellcolor{NYCband}\textbf{NYC}} &  
    \multicolumn{4}{c}{\cellcolor{TKYband}\textbf{TKY}} &  
    \multicolumn{4}{c}{\cellcolor{CAband}\textbf{CA}} \\   
    
    \cmidrule(lr){2-5}\cmidrule(lr){6-9}\cmidrule(lr){10-13}
    
    & \hNYC{ND@1} & \hNYC{ND@5} & \hNYC{ND@10} & \hNYC{MRR}
    & \hTKY{ND@1} & \hTKY{ND@5} & \hTKY{ND@10} & \hTKY{MRR}
    & \hCA{ND@1}  & \hCA{ND@5}  & \hCA{ND@10}  & \hCA{MRR} \\
    \midrule

    LSTM         & 0.1306 & 0.2336 & 0.2585 & 0.2259
                 & 0.1110 & 0.2233 & 0.2496 & 0.1952
                 & 0.0864 & 0.1459 & 0.1711 & 0.1554 \\
    PLSPL        & 0.1601 & 0.3048 & 0.3336 & 0.2849
                 & 0.1495 & 0.2831 & 0.3143 & 0.2642
                 & 0.1084 & 0.1759 & 0.2029 & 0.1678 \\
    HME          & 0.1619 & 0.2806 & 0.3226 & 0.2787
                 & 0.1535 & 0.2637 & 0.2924 & 0.2366
                 & 0.1181 & 0.1886 & 0.2232 & 0.1945 \\
    GETNext      & 0.2244 & 0.3736 & 0.4046 & 0.3472
                 & 0.1767 & 0.3072 & 0.3297 & 0.2934
                 & 0.1342 & 0.2188 & 0.2468 & 0.2121 \\
    AGRAN        & 0.2202 & 0.3638 & 0.3792 & 0.3343
                 & 0.1755 & 0.2989 & 0.3261 & 0.2879
                 & 0.1329 & 0.2121 & 0.2331 & 0.2043 \\
    MCLP         & \second{0.2404} & 0.3674 & 0.3973 & 0.3507
                 & 0.1662 & 0.3110 & 0.3415 & 0.3199
                 & 0.1324 & 0.1914 & 0.2121 & 0.1895 \\
    GeoMamba'24 & 0.1988 & 0.3392 & 0.3506 & 0.3246
                 & 0.1851 & 0.2953 & 0.3205 & 0.2858
                 & 0.1256 & 0.2029 & 0.2215 & 0.1962 \\
    GeoMamba'25 & 0.2377 & \second{0.3786} & 0.4012 & \second{0.3566}
                 & \second{0.2157} & \second{0.3402} & 0.3686 & 0.3209
                 & 0.1388 & 0.2485 & 0.2754 & 0.2373 \\
    $\mathrm{HMamba}_{\mathrm{full}}$  & 0.2204 & 0.3679 & 0.4031 & 0.3465
                 & 0.1828 & 0.3341 & 0.3673 & 0.3127
                 & 0.1366 & \second{0.2501} & \second{0.2792} & \second{0.2421} \\
    $\mathrm{HMamba}_{\mathrm{half}}$  & 0.1896 & 0.3453 & 0.3767 & 0.3222
                 & 0.1945 & 0.3295 & 0.3603 & 0.3118
                 & \second{0.1423} & 0.2381 & 0.2648 & 0.2317 \\
    HMST         & 0.2138 & 0.3747 & \second{0.4063} & 0.3482
                 & 0.1925 & 0.3325 & \second{0.3690} & \second{0.3257}
                 & 0.1356 & 0.2325 & 0.2680 & 0.2300 \\
    HVGAE        & 0.2271 & 0.3651 & 0.3982 & 0.3470
                 & 0.1977 & 0.3167 & 0.3455 & 0.3180
                 & 0.1391 & 0.2325 & 0.2658 & 0.2367 \\

    \midrule
    \cellcolor{black!4}\textbf{GTR-Mamba} &
    \best{0.2473} & \best{0.3949} & \best{0.4266} & \best{0.3709}
    & \best{0.2502} & \best{0.3757} & \best{0.4045} & \best{0.3596}
    & \best{0.1610} & \best{0.2627} & \best{0.2865} & \best{0.2548} \\
    \rowcolor{black!2}
    \textit{improv.} &
    \multicolumn{1}{c}{\up{+2.87\%}} & \multicolumn{1}{c}{\up{+4.31\%}} & \multicolumn{1}{c}{\up{+5.00\%}} & \multicolumn{1}{c}{\up{+3.93\%}}
    & \multicolumn{1}{c}{\up{+15.99\%}} & \multicolumn{1}{c}{\up{+10.44\%}} & \multicolumn{1}{c}{\up{+9.62\%}} & \multicolumn{1}{c}{\up{+10.41\%}}
    & \multicolumn{1}{c}{\up{+13.14\%}} & \multicolumn{1}{c}{\up{+5.04\%}} & \multicolumn{1}{c}{\up{+2.61\%}} & \multicolumn{1}{c}{\up{+5.25\%}} \\
    \bottomrule
    \end{tabular}%
}
\label{tab:performance_metrics}
\end{table*}
\subsection{Baseline Model}

We compare GTR-Mamba against 12 baselines, categorized into Euclidean and Hyperbolic methods.
\textbf{Euclidean-based methods} include:
\textbf{LSTM}~\cite{hochreiter1997long} and \textbf{PLSPL}~\cite{wu2020personalized}, which utilize RNNs and attention for sequential preference learning;
\textbf{GETNext}~\cite{yang2022getnext}, which incorporates global trajectory flow maps;
\textbf{AGRAN}~\cite{wang2023adaptive}, employing adaptive graph structure learning;
\textbf{MCLP}~\cite{sun2024going}, utilizing topic modeling and multi-head attention;
\textbf{GeoMamba}~\cite{chen2024geomamba}, Leveraging Mamba's linear complexity with a hierarchical geography encoder for efficient sequential modeling and 
\textbf{GeoMamba}~\cite{qin2025geomamba}, extending SSMs with a GaPPO operator to model multi-granular spatio-temporal transitions;
\textbf{Hyperbolic-based methods} include:
\textbf{HME}~\cite{feng2020hme}, embedding check-ins into a Poincaré ball;
\textbf{HMST}~\cite{qiao2025hyperbolic}, employing hyperbolic rotations to jointly model hierarchical structures and multi-semantic transitions;
\textbf{HVGAE}~\cite{liu2025hyperbolic}, integrating a Hyperbolic GCN and Variational Graph Auto-Encoder with Rotary Position Mamba to capture hierarchical and sequential POI relationships; and 
\textbf{HMamba}~\cite{zhang2025hmamba}, combining Mamba's efficiency with hyperbolic geometry. We evaluate both the \textit{Full} version (curvature-aware state spaces) and the \textit{Half} version (simplified implementation). As HMamba was originally designed for standard sequential recommendation, we adapted it for POI tasks by incorporating temporal and geographical coordinate encoders. 

To distinguish between the two identically named GeoMamba models, we append their year of publication. The two variants of HMamba are differentiated by the subscripts "full" and "half," respectively. Furthermore, as HMamba was originally designed for sequential recommendation, we have adapted it for our POI recommendation task to ensure a fairer comparison by incorporating encodings for temporal granularity and geographical coordinates to integrate spatio-temporal information.

\subsection{Performance Comparison With Baselines}
We first compare our model with 12 baseline models (Table \ref{tab:performance_metrics}). Our model consistently outperforms all baselines on the NYC and TKY (Foursquare) and CA (Gowalla) datasets. Improvements range from 2.61\% to 15.99\% in NDCG and 3.93\% to 10.41\% in MRR, demonstrating robustness in capturing spatio-temporal patterns.

Among baselines, Transformer-based approaches (GETNext and MCLP) surpass traditional models (LSTM, PLSPL) via multi-head attention mechanisms that capture complex sequential dependencies. AGRAN improves upon traditional GCNs through adaptive graph structures, while GeoMamba balances accuracy and efficiency using geographical encoding and Mamba's sequential updating. However, these Euclidean-based models are constrained. In contrast, our model operates in hyperbolic space to comprehensively capture hierarchical patterns, leveraging Mamba for efficient, stable updates.
\begin{table}[ht]
\captionsetup{aboveskip=0pt, belowskip=3pt}
\centering
\caption{Performance comparison of model variants.}
\setlength{\tabcolsep}{4pt}
\renewcommand{\arraystretch}{1.18}
\small
\resizebox{0.48\textwidth}{!}{
\begin{tabular}{l *{3}{Y} *{3}{C}}
\toprule
\multirow{2}{*}{\textbf{Model}} &
\multicolumn{3}{c}{\cellcolor{NYCband}\textbf{NYC}} & 
\multicolumn{3}{c}{\cellcolor{CAband}\textbf{CA}} \\  

\cmidrule(lr){2-4}\cmidrule(lr){5-7}

& \hNYC{ND@1} & \hNYC{ND@5} & \hNYC{MRR}
& \hCA{ND@1}  & \hCA{ND@5}  & \hCA{MRR} \\
\midrule

w/o SSM \rmv     & 0.1965 & 0.2928 & 0.2687 & 0.1236 & 0.2184 & 0.2027 \\
w/o HE  \rmv     & 0.2143 & 0.3076 & 0.2916 & 0.1398 & 0.2464 & 0.2355 \\
w/o HB  \rmv     & 0.2308 & 0.3789 & 0.3497 & 0.1432 & 0.2387 & 0.2371 \\
w/o STC \rmv     & 0.2374 & 0.3702 & 0.3544 & 0.1457 & 0.2406 & 0.2365 \\
Fixed-Base \rmv  & 0.2394 & 0.3811 & 0.3610 & 0.1508 & 0.2434 & 0.2353 \\
\midrule
\cellcolor{FullRow}\textbf{Full model} &
\bestab{0.2473} & \bestab{0.3949} & \bestab{0.3709} &
\bestab{0.1610} & \bestab{0.2627} & \bestab{0.2548} \\
\bottomrule
\end{tabular}
}
\label{ablation_results}
\end{table}
Secondly, our method significantly outperforms other hyperbolic approaches. HME is limited to node embeddings, while HVGAE and HMST model relationships better but lack powerful sequential update mechanisms. HMamba performs well but suffers computational overhead from direct manifold updates and lacks context handling. Conversely, our model's geometry-to-tangent-space pathway ensures stable computation, and our novel exogenously-driven SSM provides higher accuracy in complex scenarios. Overall, our hyperbolic Mamba approach outperforms all baselines across the three datasets.

\subsection{Ablation Study}
We conducted a comprehensive ablation study to validate the effectiveness of the individual components within our proposed model. Specifically, we designed the following ablation settings:
\begin{itemize}
\item w/o SSM (without State Space Model): This variant removes the geometry-to-tangent-space routing Mamba layer.
\item w/o HE (without Hyperbolic Embedding): This version removes the initial hyperbolic embeddings that are learned with rotation.
\item w/o HB (without Hyperbolic space): This variant implements the GTR-Mamba framework entirely in Euclidean space.
\item w/o STC (without Spatio-Temporal Channel): This variant removes the spatio-temporal encoding channel.
\item Fixed-Base: This variant performs all updates in a single global tangent space.
\end{itemize}

The ablation study results are reported in Table \ref{ablation_results}, presenting the ND@1, ND@5, and MRR metrics. Using the complete GTR-Mamba model as the baseline, we derive the following observations. The most significant performance degradation across all metrics is observed when the entire Mamba sequence modeling module is removed (w/o SSM), underscoring the critical role of dynamic relational modeling in user trajectories for personalized POI recommendation. The second most impactful variant is the one without initialized embeddings (w/o HE), highlighting the importance of rotation-based modeling for capturing complex semantic relationships. The variant excluding hyperbolic space (w/o HB) also exhibits a notable decline in all three metrics, as Euclidean space fails to effectively capture latent hierarchical structures, demonstrating the advantage of hyperbolic geometry. The variant without Euclidean spatiotemporal information (w/o STC) indicates that Euclidean encodings provide complementary support to the hyperbolic model. Removing Parallel Transport (Fixed-Base) leads to a decline in all metrics, suggesting that the PT mechanism effectively mitigates the distortion caused by tangent space approximation.

\begin{figure}[t]
\centering
\includegraphics[width=1\linewidth]{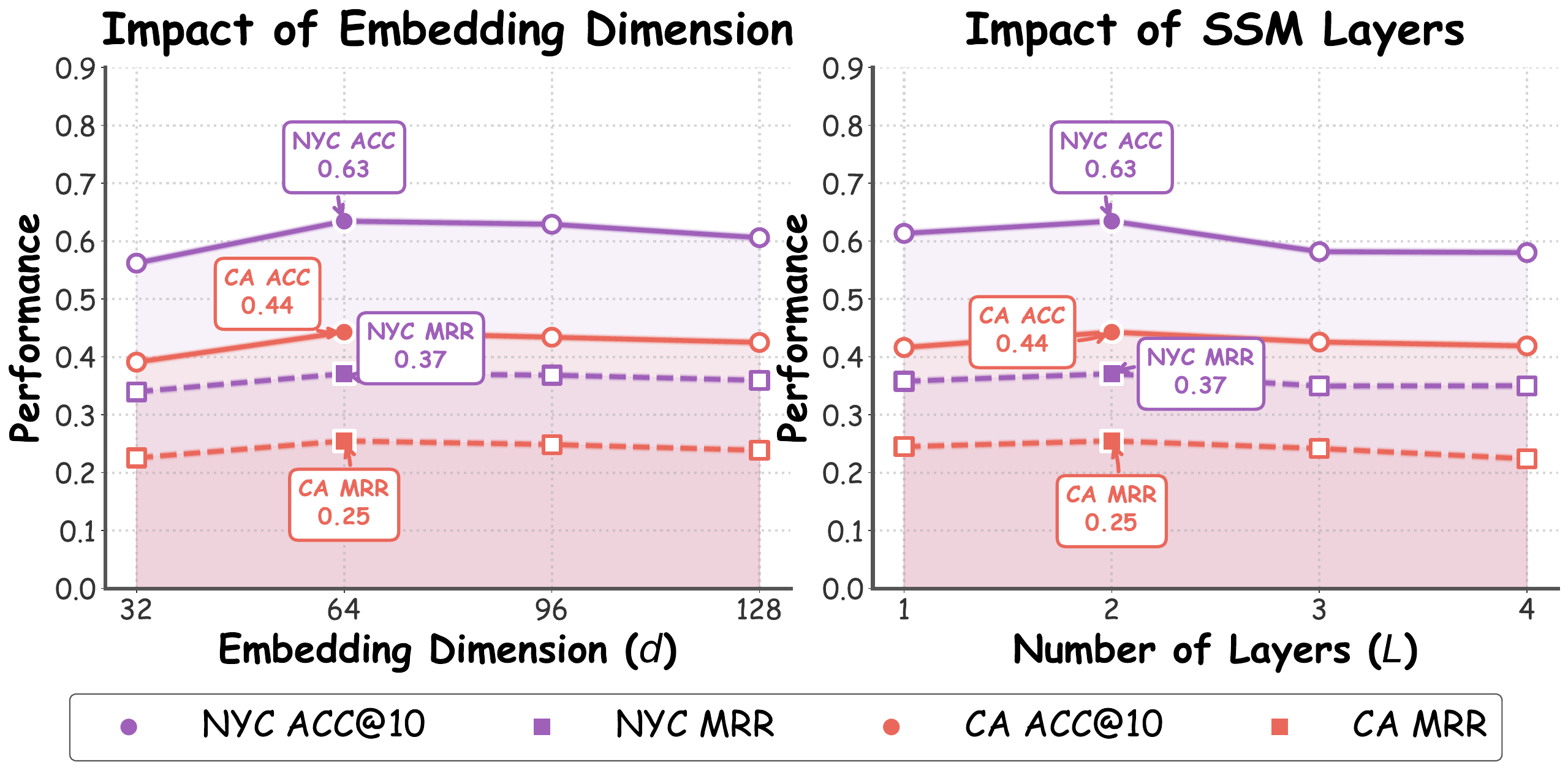}
\caption{Performance w.r.t. different embedding dimensions and SSM layers}
\Description[none]{}
\label{sensi}
\vspace{-0.4cm} 
\end{figure}

\begin{figure}[t]

\centering
\includegraphics[width=1\linewidth]{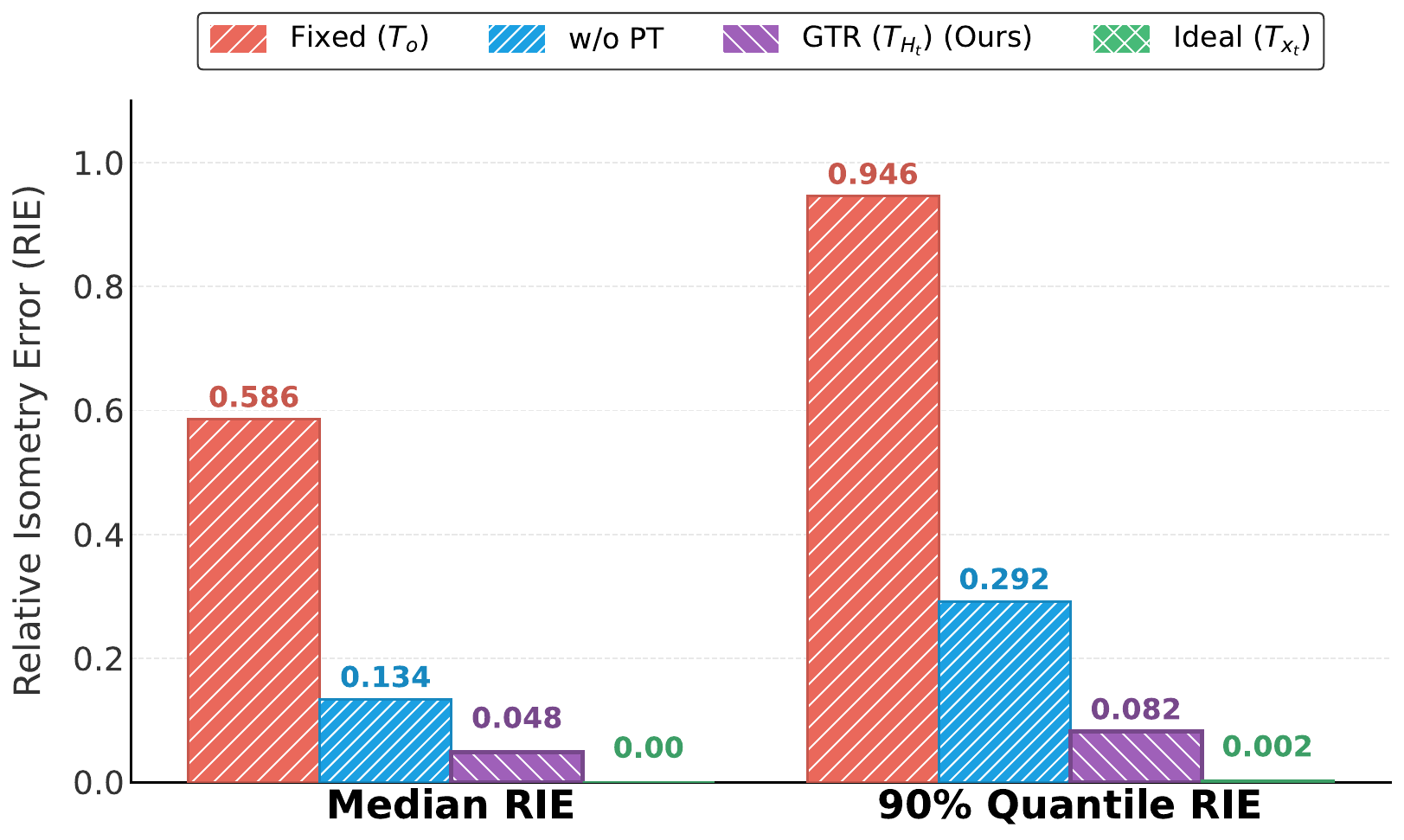}
\caption{Median and 90\% Quantile of RIE. \textbf{GTR} (purple) minimizes distortion. \textbf{w/o PT} (blue) underperforms GTR due to geometric misalignment, while \textbf{Fixed} (red) suffers from collapse.}
\Description[none]{}
\label{fig:distortion_cdf}
\end{figure}

\subsection{Sensitivity Analysis}
We analyze the impact of key hyperparameters on GTR-Mamba, specifically the embedding dimension and the number of SSM layers, with results summarized in Figure \ref{sensi}. Crucially, since the tangent space coordinates of hyperbolic embeddings serve directly as state vectors for the Selective SSM, the SSM state dimension is inherently coupled with the node embedding dimension.

\subsubsection{Embedding Dimension}
We evaluate embedding dimensions $d \in \{32, 64, 96, 128\}$ on the NYC and CA datasets. Note that for this analysis, we report ACC@10 and MRR, distinct from the NDCG metrics used in the main experiments. As shown in Figure \ref{sensi} (Left), performance peaks at $d=64$ and degrades with higher dimensions. Consequently, we set $d=64$ for all experiments.

\subsubsection{Number of SSM Layers}
We examine the effect of model depth by varying the number of SSM layers $L \in \{1, 2, 3, 4\}$. Figure \ref{sensi} (Right) indicates that while deeper models capture higher-order spatiotemporal interactions, performance plateaus beyond $L=2$ while computational costs (time and memory) rise significantly. We therefore adopt a 2-layer configuration to balance efficiency and accuracy.

\subsection{Geometric Distortion Analysis}
\label{sec:distortion_analysis}
To quantify the approximation error, we use the \textit{Relative Isometry Error} (RIE). For consecutive POIs $(x_t, x_{t+1})$, RIE is defined as $| \hat{d} - d_{\mathbb{H}} | / (d_{\mathbb{H}} + \varepsilon)$, where $d_{\mathbb{H}}$ is the geodesic distance and $\hat{d}$ is the tangent space distance. We compare four settings: (1) \textbf{Fixed ($T_o$)}: projecting to the origin; (2) \textbf{w/o PT}: moving base without parallel transport; (3) \textbf{GTR ($T_{H_t}$)}: our full model anchored at $H_t$; and (4) \textbf{Ideal ($T_{x_t}$)}: the theoretical lower bound.

\begin{table}[t]
\captionsetup{aboveskip=0pt, belowskip=2pt}
\centering
\caption{Time complexity analysis of GTR-Mamba components.}
\setlength{\tabcolsep}{6pt}
\renewcommand{\arraystretch}{1.15}
\begin{tabular}{l l}
\toprule
\textbf{Component} & \textbf{Time Complexity} \\
\midrule
Hyperbolic Representation Learning & $O(|\mathcal{E}_s|\cdot (K{+}1)\cdot d)$ \\
Spatio-temporal Channel (RFF) & $O(L\cdot m + L\cdot d)$ \\
GTR-Mamba Layer & $O(n\cdot L\cdot d^2)$ \\
\bottomrule
\end{tabular}
\label{tab:complexity}
\end{table}

\begin{figure*}
\captionsetup{skip=3pt,belowskip=-15pt}
 \centering
 \includegraphics[width=\textwidth]{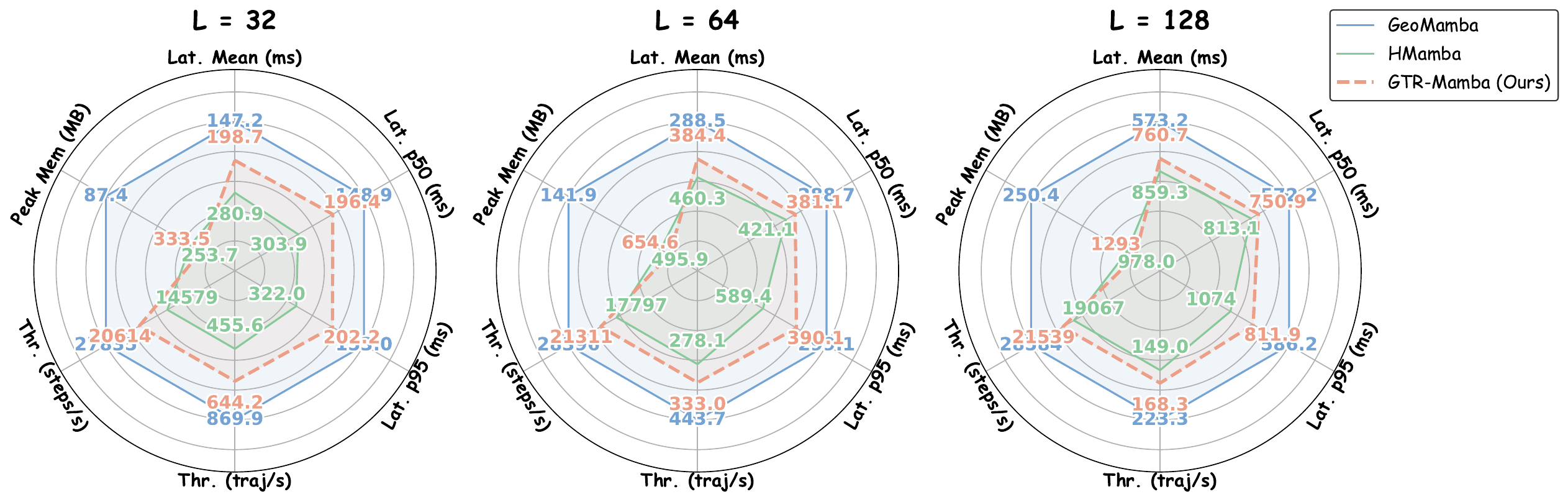}
\caption{Efficiency comparison: End-to-end inference latency (ms/batch), throughput (steps/s) and peak GPU memory (MB). GTR-Mamba (red) is slower than Euclidean GeoMamba variants (blue) due to necessary hyperbolic primitives, but remains faster than fully hyperbolic HMamba (green) by routing the scan core to the tangent space and avoiding heavy manifold recurrences.}
 \label{inference_efficiency}
 \Description[none]{}
\end{figure*}
\begin{figure}[t]
\centering
\includegraphics[width=1\linewidth]{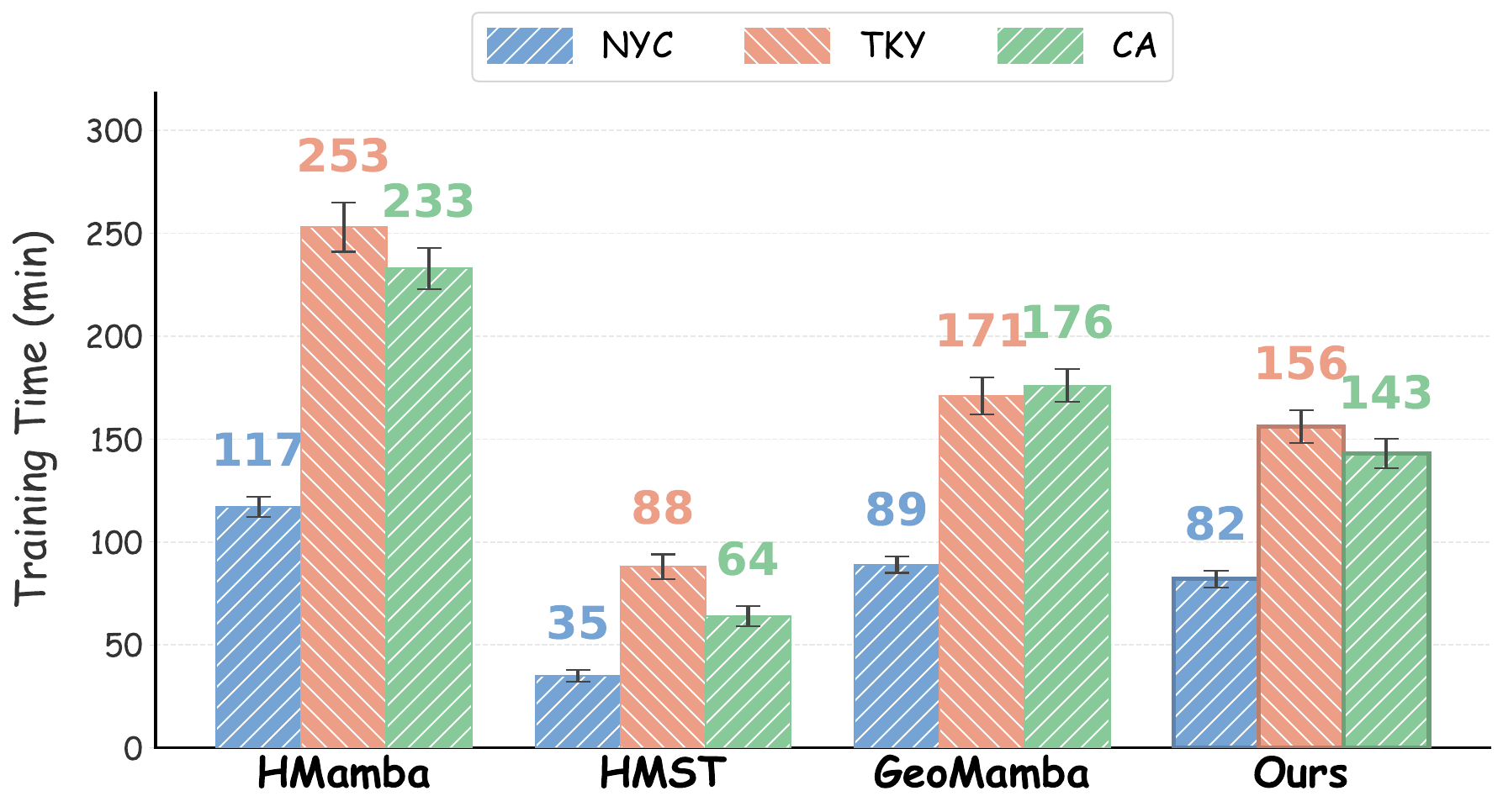}
\caption{Training efficiency analysis. The curves depict the wall-clock time required for each model to reach 95\% of its optimal MRR convergence. GTR-Mamba achieves a superior balance, converging faster than complex methods while reaching higher performance ceilings.}
\Description[none]{}
\label{time}
\vspace{-0.4cm} 
\end{figure}

Figure~\ref{fig:distortion_cdf} reports the Median and 90\% Quantile of RIE on the NYC dataset. The \textbf{Fixed} strategy (red) shows severe distortion, confirming that projecting boundary nodes to a fixed origin induces severe metric distortion, especially for points far from the origin. The \textbf{w/o PT} variant (blue) reduces error but still lags behind GTR, indicating that moving the base without geometric calibration causes hidden state misalignment. Crucially, \textbf{GTR} (purple) achieves minimal error (median 0.048), substantially closer to the \textbf{Ideal} bound (green). This verifies that Parallel Transport is essential for maintaining metric fidelity in a dynamic tangent space.

\subsection{Efficiency}
Our GTR-Mamba consists of three components: (i) hyperbolic representation learning on the Lorentz manifold, (ii) a Euclidean spatio-temporal channel, and (iii) the core GTR-Mamba layer with tangent-space selective scanning. We evaluate efficiency from three aspects: asymptotic complexity, end-to-end training time-to-quality, and end-to-end inference latency/throughput.

\subsubsection{Asymptotic complexity}
We analyze the asymptotic time complexity of each component below.

\paragraph{Variable descriptions.}
\begin{itemize}
    \item \textbf{Hyperbolic Representation Learning:} $|V|$ is the number of entities (users, POIs, categories, regions), $d$ is the embedding dimension, $|\mathcal{E}_s|$ is the number of sampled positive edges used to compute $\mathcal{L}_{\text{rot}}$ (e.g., per epoch or per mini-batch), and $K$ is the number of negative samples per positive edge.
    \item \textbf{Spatio-temporal Channel:} $L$ is the sequence length, $m$ is the total number of Random Fourier Features (including multi-scale bases), and $d$ is the model dimension after projection.
    \item \textbf{GTR-Mamba Layer:} $n$ is the number of stacked layers. The $O(d^2)$ term comes from per-position linear projections for context-driven parameterization (e.g., $A_{\text{proj}}, B_{\text{proj}}, C_{\text{proj}}$). The $O(d)$ term corresponds to the linear-time selective scan with diagonal $\mathbf{A}$, plus Lorentz manifold primitives (log/exp maps, inner products, and closed-form PT), each costing $O(d)$ per position. We treat the SSM state size and Euclidean context width as $O(d)$ in our implementation, so the overall per-layer cost remains quadratic in $d$.
\end{itemize}

\subsubsection{End-to-end inference latency and throughput}
We benchmark end-to-end inference on a single GPU with FP16 AMP. Unless otherwise stated, we use batch size $B{=}128$, warmup 20 iterations, and report statistics over 200 timed iterations. Latency is measured per batch (ms) and we report mean/p50/p95. Throughput is reported in trajectories/s and steps/s, and peak GPU memory is reported in MB.

As shown in Figure~\ref{inference_efficiency}, GTR-Mamba sits between Euclidean and fully-hyperbolic SSM baselines in end-to-end efficiency. Compared to GeoMamba \cite{qin2025geomamba}, GTR-Mamba incurs a moderate constant-factor overhead (mean latency $1.33$--$1.35\times$, throughput $0.74$--$0.75\times$), attributable to manifold-related operations and geometric scoring. Meanwhile, GTR-Mamba remains faster than HMamba across sequence lengths (throughput $1.13$--$1.41\times$ higher), while using moderately higher peak memory ($\sim 1.31\times$).

\subsubsection{Training time-to-quality}
Additionally, we compare the training time of our model against baseline models across three datasets. For a fair comparison, we measure the total wall-clock time required for each model to reach a fixed quality threshold under identical experimental conditions. Following prior practice, we define convergence as achieving $95\%$ of the optimal MRR of each method. As shown in Figure~\ref{time}, our model achieves a favorable balance between time efficiency and ranking performance. GeoMamba referenced here corresponds to GeoMamba'25~\cite{qin2025geomamba}, as distinguished from GeoMamba'24~\cite{chen2024geomamba} in baseline comparisons.

\section{Conclusion}
\label{sec:c}
In this paper, we propose GTR-Mamba, a novel framework that effectively reconciles the representational superiority of hyperbolic geometry with the computational efficiency of Selective State Space Models for next POI recommendation. GTR-Mamba executes high-order sequence modeling within the computationally efficient Euclidean tangent space, while employing a Parallel Transport (PT) mechanism to dynamically align tangent spaces across varying reference points. This effectively mitigates the distortion arising from tangent space approximation and preserves geometric consistency. This process is coordinated by an exogenous spatio-temporal channel, enabling the internal dynamics of the SSM to be explicitly driven by irregular time intervals and geographical contexts. Extensive empirical evaluations on three real-world LBSN datasets demonstrate that GTR-Mamba consistently outperforms state-of-the-art baselines, including advanced graph-based and Mamba-based approaches. Our ablation studies and efficiency analysis further validate that the framework achieves a superior trade-off between recommendation accuracy and training scalability. For future work, we plan to extend this geometry-aware mechanism to other Riemannian manifolds, such as spherical or product spaces.

\bibliographystyle{ACM-Reference-Format}
\bibliography{main}

@String{Computing = "Computing" }

@String{Computer = "{IEEE} Computer" }

@article{wang2019sequential,
  title={Sequential recommender systems: challenges, progress and prospects},
  author={Wang, Shoujin and Hu, Liang and Wang, Yan and Cao, Longbing and Sheng, Quan Z and Orgun, Mehmet},
  journal={arXiv preprint arXiv:2001.04830},
  year={2019}
}

@article{yang2014modeling,
  title={Modeling user activity preference by leveraging user spatial temporal characteristics in LBSNs},
  author={Yang, Dingqi and Zhang, Daqing and Zheng, Vincent W and Yu, Zhiyong},
  journal={IEEE Transactions on Systems, Man, and Cybernetics: Systems},
  volume={45},
  number={1},
  pages={129--142},
  year={2014},
  publisher={IEEE}
}

@article{sanchez2022point,
  title={Point-of-interest recommender systems based on location-based social networks: a survey from an experimental perspective},
  author={S{\'a}nchez, Pablo and Bellog{\'\i}n, Alejandro},
  journal={ACM Computing Surveys (CSUR)},
  volume={54},
  number={11s},
  pages={1--37},
  year={2022},
  publisher={ACM New York, NY}
}

@article{baral2018caps,
  title={CAPS: Context aware personalized POI sequence recommender system},
  author={Baral, Ramesh and Li, Tao and Zhu, XiaoLong},
  journal={arXiv preprint arXiv:1803.01245},
  year={2018}
}

@inproceedings{yuan2013time,
  title={Time-aware point-of-interest recommendation},
  author={Yuan, Quan and Cong, Gao and Ma, Zongyang and Sun, Aixin and Thalmann, Nadia Magnenat-},
  booktitle={Proceedings of the 36th international ACM SIGIR conference on Research and development in information retrieval},
  pages={363--372},
  year={2013}
}

@inproceedings{law2019lorentzian,
  title={Lorentzian distance learning for hyperbolic representations},
  author={Law, Marc and Liao, Renjie and Snell, Jake and Zemel, Richard},
  booktitle={International Conference on Machine Learning},
  pages={3672--3681},
  year={2019},
  organization={PMLR}
}

@inproceedings{lian2020geography,
  title={Geography-aware sequential location recommendation},
  author={Lian, Defu and Wu, Yongji and Ge, Yong and Xie, Xing and Chen, Enhong},
  booktitle={Proceedings of the 26th ACM SIGKDD international conference on knowledge discovery \& data mining},
  pages={2009--2019},
  year={2020}
}

@inproceedings{lim2022hierarchical,
  title={Hierarchical multi-task graph recurrent network for next poi recommendation},
  author={Lim, Nicholas and Hooi, Bryan and Ng, See-Kiong and Goh, Yong Liang and Weng, Renrong and Tan, Rui},
  booktitle={Proceedings of the 45th international ACM SIGIR conference on Research and development in Information Retrieval},
  pages={1133--1143},
  year={2022}
}

@article{chami2019hyperbolic,
  title={Hyperbolic graph convolutional neural networks},
  author={Chami, Ines and Ying, Zhitao and R{\'e}, Christopher and Leskovec, Jure},
  journal={Advances in neural information processing systems},
  volume={32},
  year={2019}
}

@inproceedings{yang2022hicf,
  title={Hicf: Hyperbolic informative collaborative filtering},
  author={Yang, Menglin and Li, Zhihao and Zhou, Min and Liu, Jiahong and King, Irwin},
  booktitle={Proceedings of the 28th ACM SIGKDD Conference on Knowledge Discovery and Data Mining},
  pages={2212--2221},
  year={2022}
}

@article{guo2023hyperbolic,
  title={Hyperbolic contrastive graph representation learning for session-based recommendation},
  author={Guo, Naicheng and Liu, Xiaolei and Li, Shaoshuai and Ha, Mingming and Ma, Qiongxu and Wang, Binfeng and Zhao, Yunan and Chen, Linxun and Guo, Xiaobo},
  journal={IEEE Transactions on Knowledge and Data Engineering},
  year={2023},
  publisher={IEEE}
}

@article{li2022hyperbolic,
  title={Hyperbolic neural collaborative recommender},
  author={Li, Anchen and Yang, Bo and Huo, Huan and Chen, Hongxu and Xu, Guandong and Wang, Zhen},
  journal={IEEE Transactions on Knowledge and Data Engineering},
  volume={35},
  number={9},
  pages={9114--9127},
  year={2022},
  publisher={IEEE}
}

@inproceedings{wang2023hdnr,
  title={Hdnr: A hyperbolic-based debiased approach for personalized news recommendation},
  author={Wang, Shicheng and Guo, Shu and Wang, Lihong and Liu, Tingwen and Xu, Hongbo},
  booktitle={Proceedings of the 46th International ACM SIGIR Conference on Research and Development in Information Retrieval},
  pages={259--268},
  year={2023}
}

@article{wang2021hypersorec,
  title={Hypersorec: Exploiting hyperbolic user and item representations with multiple aspects for social-aware recommendation},
  author={Wang, Hao and Lian, Defu and Tong, Hanghang and Liu, Qi and Huang, Zhenya and Chen, Enhong},
  journal={ACM Transactions on Information Systems (TOIS)},
  volume={40},
  number={2},
  pages={1--28},
  year={2021},
  publisher={ACM New York, NY}
}

@inproceedings{bose2020latent,
  title={Latent variable modelling with hyperbolic normalizing flows},
  author={Bose, Joey and Smofsky, Ariella and Liao, Renjie and Panangaden, Prakash and Hamilton, Will},
  booktitle={International conference on machine learning},
  pages={1045--1055},
  year={2020},
  organization={PMLR}
}

@inproceedings{shimizu2024fashion,
  title={A fashion item recommendation model in hyperbolic space},
  author={Shimizu, Ryotaro and Wang, Yu and Kimura, Masanari and Hirakawa, Yuki and Wada, Takashi and Saito, Yuki and McAuley, Julian},
  booktitle={Proceedings of the IEEE/CVF Conference on Computer Vision and Pattern Recognition},
  pages={8377--8383},
  year={2024}
}

@article{zhang2004principal,
  title={Principal manifolds and nonlinear dimensionality reduction via tangent space alignment},
  author={Zhang, Zhenyue and Zha, Hongyuan},
  journal={SIAM journal on scientific computing},
  volume={26},
  number={1},
  pages={313--338},
  year={2004},
  publisher={SIAM}
}

@article{liu2025hybrid,
  title={Hybrid-Emba3D: Geometry-Aware and Cross-Path Feature Hybrid Enhanced State Space Model for Point Cloud Classification},
  author={Liu, Bin and Wang, Chunyang and Liu, Xuelian and Xi, Guan and Zhang, Ge and Yao, Ziteng and Dong, Mengxue},
  journal={arXiv e-prints},
  pages={arXiv--2505},
  year={2025}
}

@article{patil2025hierarchical,
  title={Hierarchical Mamba Meets Hyperbolic Geometry: A New Paradigm for Structured Language Embeddings},
  author={Patil, Sarang and Pandey, Ashish Parmanand and Koutis, Ioannis and Xu, Mengjia},
  journal={arXiv preprint arXiv:2505.18973},
  year={2025}
}

@article{yang2023hyperbolic,
  title={Hyperbolic graph learning for social recommendation},
  author={Yang, Yonghui and Wu, Le and Zhang, Kun and Hong, Richang and Zhou, Hailin and Zhang, Zhiqiang and Zhou, Jun and Wang, Meng},
  journal={IEEE Transactions on Knowledge and Data Engineering},
  volume={36},
  number={12},
  pages={8488--8501},
  year={2023},
  publisher={IEEE}
}

@inproceedings{yang2024hypformer,
  title={Hypformer: Exploring efficient transformer fully in hyperbolic space},
  author={Yang, Menglin and Verma, Harshit and Zhang, Delvin Ce and Liu, Jiahong and King, Irwin and Ying, Rex},
  booktitle={Proceedings of the 30th ACM SIGKDD Conference on Knowledge Discovery and Data Mining},
  pages={3770--3781},
  year={2024},
  publisher={ACM}
}

@inproceedings{chen2022modeling,
  title={Modeling scale-free graphs with hyperbolic geometry for knowledge-aware recommendation},
  author={Chen, Yankai and Yang, Menglin and Zhang, Yingxue and Zhao, Mengchen and Meng, Ziqiao and Hao, Jianye and King, Irwin},
  booktitle={Proceedings of the fifteenth ACM international conference on web search and data mining},
  pages={94--102},
  year={2022}
}

@inproceedings{du2022hakg,
  title={HAKG: Hierarchy-aware knowledge gated network for recommendation},
  author={Du, Yuntao and Zhu, Xinjun and Chen, Lu and Zheng, Baihua and Gao, Yunjun},
  booktitle={Proceedings of the 45th international ACM SIGIR conference on Research and development in Information Retrieval},
  pages={1390--1400},
  year={2022}
}

@article{xie2023hierarchical,
  title={Hierarchical transformer with spatio-temporal context aggregation for next point-of-interest recommendation},
  author={Xie, Jiayi and Chen, Zhenzhong},
  journal={ACM Transactions on Information Systems},
  volume={42},
  number={2},
  pages={1--30},
  year={2023},
  publisher={ACM New York, NY, USA}
}

@inproceedings{sun2021hgcf,
  title={Hgcf: Hyperbolic graph convolution networks for collaborative filtering},
  author={Sun, Jianing and Cheng, Zhaoyue and Zuberi, Saba and P{\'e}rez, Felipe and Volkovs, Maksims},
  booktitle={Proceedings of the web conference 2021},
  pages={593--601},
  year={2021}
}

@inproceedings{yu2020category,
  title={A category-aware deep model for successive POI recommendation on sparse check-in data},
  author={Yu, Fuqiang and Cui, Lizhen and Guo, Wei and Lu, Xudong and Li, Qingzhong and Lu, Hua},
  booktitle={Proceedings of the web conference 2020},
  pages={1264--1274},
  year={2020}
}

@article{zhang2020modeling,
  title={Modeling hierarchical category transition for next POI recommendation with uncertain check-ins},
  author={Zhang, Lu and Sun, Zhu and Zhang, Jie and Kloeden, Horst and Klanner, Felix},
  journal={Information Sciences},
  volume={515},
  pages={169--190},
  year={2020},
  publisher={Elsevier}
}

@article{zang2021cha,
  title={Cha: Categorical hierarchy-based attention for next poi recommendation},
  author={Zang, Hongyu and Han, Dongcheng and Li, Xin and Wan, Zhifeng and Wang, Mingzhong},
  journal={ACM Transactions on Information Systems (TOIS)},
  volume={40},
  number={1},
  pages={1--22},
  year={2021},
  publisher={ACM New York, NY}
}

@inproceedings{gu2024mamba,
  title={Mamba: Linear-time sequence modeling with selective state spaces},
  author={Gu, Albert and Dao, Tri},
  booktitle={First Conference on Language Modeling},
  year={2024}
}

@article{yang2022hyperbolic,
  title={Hyperbolic graph neural networks: A review of methods and applications},
  author={Yang, Menglin and Zhou, Min and Li, Zhihao and Liu, Jiahong and Pan, Lujia and Xiong, Hui and King, Irwin},
  journal={arXiv preprint arXiv:2202.13852},
  year={2022}
}

@article{peng2021hyperbolic,
  title={Hyperbolic deep neural networks: A survey},
  author={Peng, Wei and Varanka, Tuomas and Mostafa, Abdelrahman and Shi, Henglin and Zhao, Guoying},
  journal={IEEE Transactions on pattern analysis and machine intelligence},
  volume={44},
  number={12},
  pages={10023--10044},
  year={2021},
  publisher={IEEE}
}

@article{ganea2018hyperbolic,
  title={Hyperbolic neural networks},
  author={Ganea, Octavian and B{\'e}cigneul, Gary and Hofmann, Thomas},
  journal={Advances in neural information processing systems},
  volume={31},
  year={2018}
}

@inproceedings{yan2023spatio,
  title={Spatio-temporal hypergraph learning for next POI recommendation},
  author={Yan, Xiaodong and Song, Tengwei and Jiao, Yifeng and He, Jianshan and Wang, Jiaotuan and Li, Ruopeng and Chu, Wei},
  booktitle={Proceedings of the 46th international ACM SIGIR conference on research and development in information retrieval},
  pages={403--412},
  year={2023}
}

@inproceedings{huang2024learning,
  title={Learning time slot preferences via mobility tree for next poi recommendation},
  author={Huang, Tianhao and Pan, Xuan and Cai, Xiangrui and Zhang, Ying and Yuan, Xiaojie},
  booktitle={Proceedings of the AAAI Conference on Artificial Intelligence},
  volume={38},
  number={8},
  pages={8535--8543},
  year={2024}
}

@inproceedings{wang2022learning,
  title={Learning graph-based disentangled representations for next POI recommendation},
  author={Wang, Zhaobo and Zhu, Yanmin and Liu, Haobing and Wang, Chunyang},
  booktitle={Proceedings of the 45th international ACM SIGIR conference on research and development in information retrieval},
  pages={1154--1163},
  year={2022}
}

@inproceedings{luo2021stan,
  title={Stan: Spatio-temporal attention network for next location recommendation},
  author={Luo, Yingtao and Liu, Qiang and Liu, Zhaocheng},
  booktitle={Proceedings of the web conference 2021},
  pages={2177--2185},
  year={2021}
}

@inproceedings{duan2023clsprec,
  title={Clsprec: Contrastive learning of long and short-term preferences for next poi recommendation},
  author={Duan, Chenghua and Fan, Wei and Zhou, Wei and Liu, Hu and Wen, Junhao},
  booktitle={Proceedings of the 32nd acm international conference on information and knowledge management},
  pages={473--482},
  year={2023}
}

@article{xue2021mobtcast,
  title={MobTCast: Leveraging auxiliary trajectory forecasting for human mobility prediction},
  author={Xue, Hao and Salim, Flora and Ren, Yongli and Oliver, Nuria},
  journal={Advances in Neural Information Processing Systems},
  volume={34},
  pages={30380--30391},
  year={2021}
}

@inproceedings{zhang2022next,
  title={Next Point-of-Interest Recommendation with Inferring Multi-step Future Preferences.},
  author={Zhang, Lu and Sun, Zhu and Wu, Ziqing and Zhang, Jie and Ong, Yew Soon and Qu, Xinghua},
  booktitle={IJCAI},
  pages={3751--3757},
  year={2022}
}

@inproceedings{liu2016predicting,
  title={Predicting the next location: A recurrent model with spatial and temporal contexts},
  author={Liu, Qiang and Wu, Shu and Wang, Liang and Tan, Tieniu},
  booktitle={Proceedings of the AAAI conference on artificial intelligence},
  volume={30},
  number={1},
  year={2016}
}

@inproceedings{feng2018deepmove,
  title={Deepmove: Predicting human mobility with attentional recurrent networks},
  author={Feng, Jie and Li, Yong and Zhang, Chao and Sun, Funing and Meng, Fanchao and Guo, Ang and Jin, Depeng},
  booktitle={Proceedings of the 2018 world wide web conference},
  pages={1459--1468},
  year={2018}
}

@inproceedings{wang2021reinforced,
  title={Reinforced imitative graph representation learning for mobile user profiling: An adversarial training perspective},
  author={Wang, Dongjie and Wang, Pengyang and Liu, Kunpeng and Zhou, Yuanchun and Hughes, Charles E and Fu, Yanjie},
  booktitle={Proceedings of the AAAI Conference on Artificial Intelligence},
  volume={35},
  number={5},
  pages={4410--4417},
  year={2021}
}

@inproceedings{sun2020go,
  title={Where to go next: Modeling long-and short-term user preferences for point-of-interest recommendation},
  author={Sun, Ke and Qian, Tieyun and Chen, Tong and Liang, Yile and Nguyen, Quoc Viet Hung and Yin, Hongzhi},
  booktitle={Proceedings of the AAAI conference on artificial intelligence},
  volume={34},
  number={01},
  pages={214--221},
  year={2020}
}

@article{wu2020personalized,
  title={Personalized long-and short-term preference learning for next POI recommendation},
  author={Wu, Yuxia and Li, Ke and Zhao, Guoshuai and Qian, Xueming},
  journal={IEEE Transactions on Knowledge and Data Engineering},
  volume={34},
  number={4},
  pages={1944--1957},
  year={2020},
  publisher={IEEE}
}

@inproceedings{sun2024going,
  title={Going where, by whom, and at what time: Next location prediction considering user preference and temporal regularity},
  author={Sun, Tianao and Fu, Ke and Huang, Weiming and Zhao, Kai and Gong, Yongshun and Chen, Meng},
  booktitle={Proceedings of the 30th ACM SIGKDD Conference on Knowledge Discovery and Data Mining},
  pages={2784--2793},
  year={2024}
}

@inproceedings{feng2020hme,
  title={Hme: A hyperbolic metric embedding approach for next-poi recommendation},
  author={Feng, Shanshan and Tran, Lucas Vinh and Cong, Gao and Chen, Lisi and Li, Jing and Li, Fan},
  booktitle={Proceedings of the 43rd International ACM SIGIR Conference on research and development in information retrieval},
  pages={1429--1438},
  year={2020}
}

@inproceedings{wang2023adaptive,
  title={Adaptive graph representation learning for next POI recommendation},
  author={Wang, Zhaobo and Zhu, Yanmin and Wang, Chunyang and Ma, Wenze and Li, Bo and Yu, Jiadi},
  booktitle={Proceedings of the 46th international ACM SIGIR conference on research and development in information retrieval},
  pages={393--402},
  year={2023}
}

@inproceedings{yang2022getnext,
  title={GETNext: Trajectory flow map enhanced transformer for next POI recommendation},
  author={Yang, Song and Liu, Jiamou and Zhao, Kaiqi},
  booktitle={Proceedings of the 45th International ACM SIGIR Conference on research and development in information retrieval},
  pages={1144--1153},
  year={2022}
}

@article{hochreiter1997long,
  title={Long short-term memory},
  author={Hochreiter, Sepp and Schmidhuber, J{\"u}rgen},
  journal={Neural computation},
  volume={9},
  number={8},
  pages={1735--1780},
  year={1997},
  publisher={MIT press}
}

@article{jiang2026hierarchical,
  title={Hierarchical long and short-term preference modeling with denoising Mamba for sequential recommendation},
  author={Jiang, Wei and Fan, Yongquan and Tang, Jing and Li, Xianyong and Du, Yajun and Wang, Xiaomin},
  journal={Information Processing \& Management},
  volume={63},
  number={2},
  pages={104425},
  year={2026},
  publisher={Elsevier}
}

@inproceedings{qiao2025hyperbolic,
  title={Hyperbolic Multi-semantic Transition for Next POI Recommendation},
  author={Qiao, Hongliang and Feng, Shanshan and Zhou, Min and Li, WenTao and Li, Fan},
  booktitle={Companion Proceedings of the ACM on Web Conference 2025},
  pages={1830--1837},
  year={2025}
}

@article{zhang2025hmamba,
  title={HMamba: Hyperbolic Mamba for Sequential Recommendation},
  author={Zhang, Qianru and Wen, Honggang and Yuan, Wei and Chen, Crystal and Yang, Menglin and Yiu, Siu-Ming and Yin, Hongzhi},
  journal={arXiv preprint arXiv:2505.09205},
  year={2025}
}

@inproceedings{liu2025hyperbolic,
  title={Hyperbolic Variational Graph Auto-Encoder for Next POI Recommendation},
  author={Liu, Yuwen and Qi, Lianyong and Mao, Xingyuan and Liu, Weiming and Wang, Fan and Xu, Xiaolong and Zhang, Xuyun and Dou, Wanchun and Zhou, Xiaokang and Beheshti, Amin},
  booktitle={Proceedings of the ACM on Web Conference 2025},
  pages={3267--3275},
  year={2025}
}

@inproceedings{xu2023revisiting,
  title={Revisiting mobility modeling with graph: A graph transformer model for next point-of-interest recommendation},
  author={Xu, Xiaohang and Suzumura, Toyotaro and Yong, Jiawei and Hanai, Masatoshi and Yang, Chuang and Kanezashi, Hiroki and Jiang, Renhe and Fukushima, Shintaro},
  booktitle={Proceedings of the 31st ACM international conference on advances in geographic information systems},
  pages={1--10},
  year={2023}
}

@inproceedings{qin2025geomamba,
  title={GeoMamba: Towards Multi-granular POI Recommendation with Geographical State Space Model},
  author={Qin, Yifang and Xie, Jiaxuan and Xiao, Zhiping and Zhang, Ming},
  booktitle={Proceedings of the AAAI Conference on Artificial Intelligence},
  volume={39},
  number={12},
  pages={12479--12487},
  year={2025}
}

@article{chen2024geomamba,
  title={GeoMamba: Towards Efficient Geography-aware Sequential POI Recommendation},
  author={Chen, Jiubing and Wang, Haoyu and Shang, Jianxin},
  journal={IEEE Access},
  year={2024},
  publisher={IEEE}
}

@inproceedings{qin2023disenpoi,
  title={DisenPOI: Disentangling sequential and geographical influence for point-of-interest recommendation},
  author={Qin, Yifang and Wang, Yifan and Sun, Fang and Ju, Wei and Hou, Xuyang and Wang, Zhe and Cheng, Jia and Lei, Jun and Zhang, Ming},
  booktitle={Proceedings of the sixteenth ACM international conference on web search and data mining},
  pages={508--516},
  year={2023}
}

@inproceedings{li2021you,
  title={You are what and where you are: Graph enhanced attention network for explainable poi recommendation},
  author={Li, Zeyu and Cheng, Wei and Xiao, Haiqi and Yu, Wenchao and Chen, Haifeng and Wang, Wei},
  booktitle={Proceedings of the 30th ACM International conference on information \& knowledge management},
  pages={3945--3954},
  year={2021}
}

@article{chen2023dynamic,
  title={Dynamic personalized POI sequence recommendation with fine-grained contexts},
  author={Chen, Jing and Jiang, Wenjun and Wu, Jie and Li, Kenli and Li, Keqin},
  journal={ACM transactions on internet technology},
  volume={23},
  number={2},
  pages={1--28},
  year={2023},
  publisher={ACM New York, NY}
}

@inproceedings{baral2018close,
  title={Close: Contextualized location sequence recommender},
  author={Baral, Ramesh and Iyengar, SS and Li, Tao and Balakrishnan, N},
  booktitle={Proceedings of the 12th ACM conference on recommender systems},
  pages={470--474},
  year={2018}
}

@inproceedings{wang2019spent,
  title={SPENT: A successive POI recommendation method using similarity-based POI embedding and recurrent neural network with temporal influence},
  author={Wang, Mu-Fan and Lu, Yi-Shu and Huang, Jiun-Long},
  booktitle={2019 IEEE International Conference on Big Data and Smart Computing (BigComp)},
  pages={1--8},
  year={2019},
  organization={IEEE}
}

@article{lu2020glr,
  title={GLR: A graph-based latent representation model for successive POI recommendation},
  author={Lu, Yi-Shu and Huang, Jiun-Long},
  journal={Future Generation Computer Systems},
  volume={102},
  pages={230--244},
  year={2020},
  publisher={Elsevier}
}

\end{document}